%% file: main.tex
\documentclass{article}
\usepackage[OT1]{fontenc}

\usepackage{iclr2027_conference,times}
\input{math_commands.tex}

\usepackage{amsmath,amssymb,mathtools}
\usepackage{booktabs,multirow,array,tabularx}
\usepackage{graphicx}
\usepackage{placeins,needspace,float}
\usepackage{xcolor}
\usepackage{xspace}
\usepackage{enumitem}
\usepackage{microtype}
\usepackage{tikz}
\usetikzlibrary{arrows.meta,positioning,fit}
\usepackage{hyperref}
\usepackage{url}
\hypersetup{
  hidelinks,
  pdftitle={CalibCache: Cache-to-Full Feature Dynamics Calibration for Diffusion Transformers},
  pdfauthor={Mingyu Liang, Dingkun Xu, Jingwei Xu}
}

\definecolor{methodblue}{HTML}{215A8E}
\definecolor{methodgreen}{HTML}{2E7D5B}
\definecolor{softblue}{HTML}{EDF5FB}
\definecolor{softgreen}{HTML}{EDF7F2}
\definecolor{softgray}{HTML}{F4F5F6}
\definecolor{draftorange}{HTML}{A65300}

\newcommand{\method}{\textsc{CalibCache}\xspace}
\newcommand{\full}{\mathrm{F}}
\newcommand{\cache}{\mathrm{C}}

\newcommand{\br}{\mathbf{r}}
\newcommand{\bv}{\mathbf{v}}
\newcommand{\ba}{\mathbf{a}}

\newcommand{\bx}{\mathbf{x}}
\newcommand{\by}{\mathbf{y}}

\newcommand{\best}[1]{\textbf{#1}}

\title{\method: Cache-to-Full Feature Dynamics Calibration for Diffusion Transformers}

\author{Mingyu Liang\textsuperscript{\ensuremath{*}},
  Dingkun Xu\textsuperscript{\ensuremath{*}},
  Jingwei Xu\textsuperscript{\ensuremath{\dagger}}\\
  \normalfont School of Computer Science, Nanjing University\\[-1pt]
  \small\normalfont
  \href{mailto:522025330055@smail.nju.edu.cn}{522025330055@smail.nju.edu.cn}
  \quad \href{mailto:502025330057@smail.nju.edu.cn}{502025330057@smail.nju.edu.cn}
  \quad \href{mailto:jingweix@nju.edu.cn}{jingweix@nju.edu.cn}\\[-1pt]
  \small\normalfont \textsuperscript{\ensuremath{*}}Equal contribution
  \quad \textsuperscript{\ensuremath{\dagger}}Corresponding author}

\iclrfinalcopy

\begin{document}
\maketitle

\input{sections/00_abstract}
\input{sections/01_introduction}
\input{sections/02_related_work}
\input{sections/03_method}
\input{sections/04_experiments}

\input{sections/05_ablation}
\FloatBarrier
\Needspace{9\baselineskip}
\input{sections/05_conclusion}
\newpage
\input{sections/06_statements}

\bibliography{paper_references}
\bibliographystyle{iclr2027_conference}

\appendix
\input{sections/07_appendix_implementation}

\end{document}

%% file: math_commands.tex
\usepackage{amsmath,amsfonts,bm}

\def\eqref#1{equation~\ref{#1}}

\def\1{\bm{1}}

\DeclareMathAlphabet{\mathsfit}{\encodingdefault}{\sfdefault}{m}{sl}
\SetMathAlphabet{\mathsfit}{bold}{\encodingdefault}{\sfdefault}{bx}{n}



%% file: sections/00_abstract.tex
\begin{abstract}
Diffusion Transformers generate high-quality images, but iterative sampling remains computationally expensive.
Feature caching reduces the number of full network evaluations,
yet its approximation error can become more severe under short sampling schedules, where fewer full-compute steps and a coarser sampling grid can make history-based extrapolation less reliable.
In this paper, we propose \method{}, a lightweight offline calibration method that corrects systematic discrepancies between cached and full-compute feature dynamics.
Specifically, \method{} calibrates feature acceleration, defined as the second-order finite difference across sampling steps.
Using only a single pair of full-compute and naive-cache trajectories at each setting, \method{} fits scalar acceleration maps offline by centered least squares.
These maps are then applied during sampling without additional full-network evaluations or online optimization.
Across FLUX.1-dev and Qwen-Image on DrawBench and DiT-XL/2 on ImageNet-256, \method{} consistently improves fidelity or generation quality over TaylorSeer under 20-step sampling at cache intervals $N=3,4,5$.
At $N=5$, it reduces FLUX.1-dev LPIPS from $0.563$ to $0.429$ while achieving a $2.86\times$ denoising speedup over full-compute sampling, and lowers DiT-XL/2 FID from $16.39$ to $10.37$ relative to TaylorSeer.
\end{abstract}

%% file: sections/01_introduction.tex
\section{Introduction}
\setlength{\textfloatsep}{12pt}
\begin{figure}[t]
  \centering
  \includegraphics[width=\linewidth]{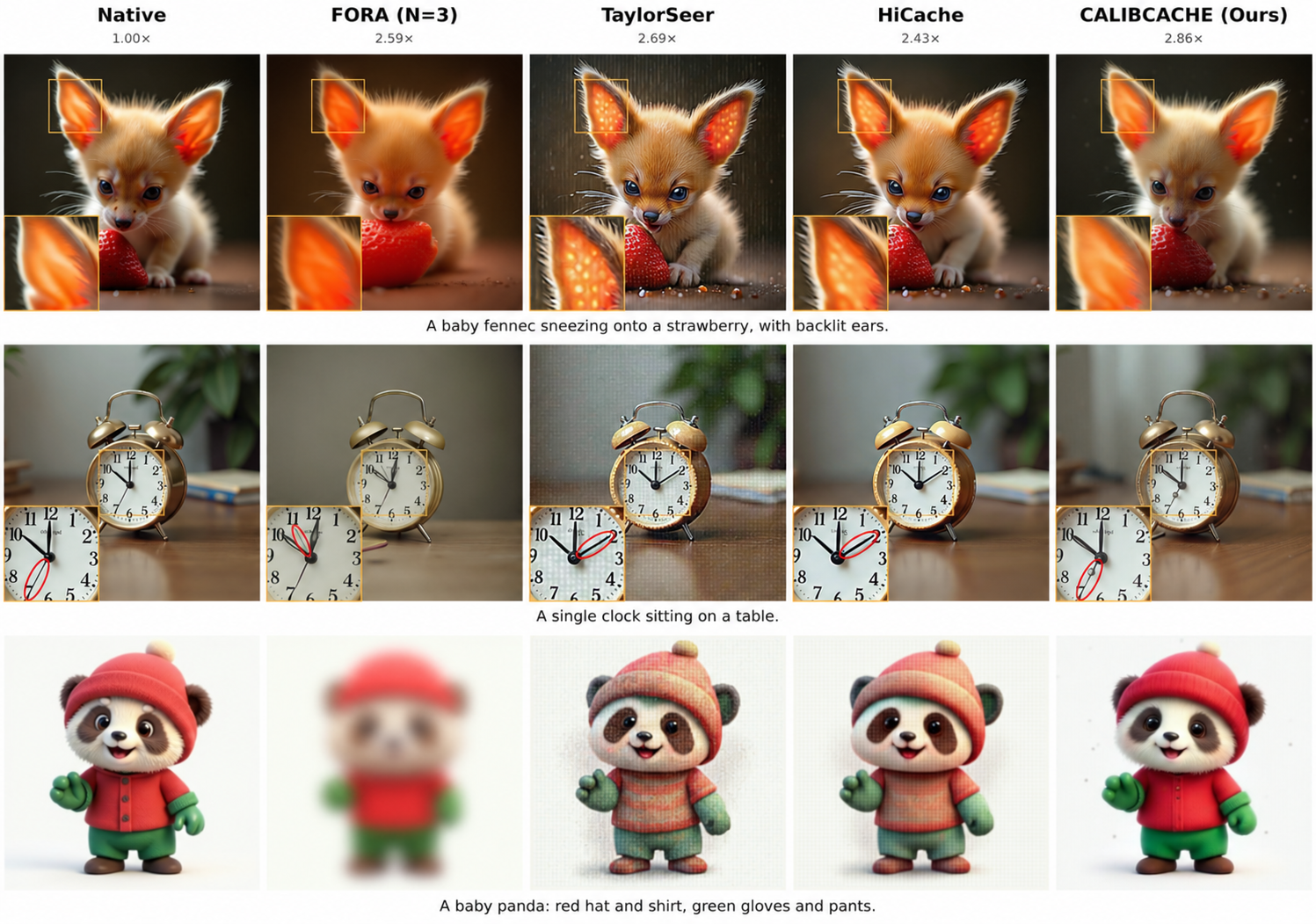}
  \caption{Five-column qualitative comparison on FLUX.1-dev with 20-step DrawBench sampling. Within each row, all methods use the same prompt and initial noise. The columns show Native, FORA ($N=3$, $2.59\times$), TaylorSeer ($N=5$, $2.69\times$), HiCache ($N=5$, $2.43\times$), and \method{} ($N=5$, $2.86\times$); speedups refer to denoising latency relative to Native. FORA uses $N=3$ to show a broadly comparable speedup. The enlarged regions highlight differences in the ear and clock-hand details; in these examples, \method{} more closely preserves the Native appearance.}
  \label{fig:intro_qualitative}
\end{figure}

Diffusion Transformers (DiTs) have become a central backbone for image generation~\citep{peebles2023dit}.
Although recent models can generate high-quality images with only a few tens of sampling steps, inference remains expensive because each step requires costly Transformer attention and feed-forward computations.
Training-free feature caching reduces this cost by reusing intermediate features across adjacent sampling steps~\citep{ma2023deepcache,selvaraju2024fora,chen2024deltadit,zou2025toca}, but reused features can deviate from their full-compute counterparts as the reuse interval grows.

Short sampling schedules can exacerbate this discrepancy. Fewer sampling steps provide fewer full-compute observations overall, while a cache interval of fixed $N$ spans a larger fraction of the rollout. History-based extrapolation can therefore become less reliable under coarse sampling, where sparse historical observations provide a weaker basis for estimating local feature dynamics and prediction errors can accumulate across reused steps. These challenges motivate a practical correction method that improves feature quality while preserving the computational savings of caching.

Existing caching methods address feature approximation in different ways.
Naive reuse is computationally inexpensive but can produce stale features at larger cache intervals.
TaylorSeer~\citep{liu2025taylorseer} and HiCache~\citep{feng2026hicache} extrapolate future features from sampling history using Taylor and scaled-Hermite predictors, respectively.
TC-Pad\'e~\citep{cui2026tcpade} uses rational residual prediction with online coefficient adaptation, while \(L^{2}P\)~\citep{shen2026l2p} learns linear combinations of historical features through a lightweight linear layer, supervised by reconstruction targets from full-compute trajectories.
Rather than learning feature evolution from full-compute trajectories alone, we ask a complementary question: \textit{can an explicitly aligned pair of full-compute and naive-cache trajectories provide direct supervision for a compact, fixed calibration of cached feature dynamics?}

In this paper, we propose \method{}, a lightweight offline method built on \emph{paired cache-to-full acceleration supervision}.
We generate one full-compute and one naive-cache trajectory from the same condition and initial noise, then pair their feature accelerations at matching sampling steps and residual boundaries.
Feature acceleration is defined as the second finite difference of residual features.
Because the two trajectories evolve independently under full computation and the prescribed cache schedule, respectively, their pairing directly exposes the cache-induced discrepancy in feature dynamics.
Centered least squares then estimates one shared scalar per step--branch group to map naive-cache acceleration to its full-compute counterpart.
The resulting supervision therefore relates feature dynamics across the two rollouts, rather than reconstructing current features from history within full-compute trajectories.

For each deployment configuration, offline preparation requires only one trajectory pair followed by closed-form scalar fitting.
The correction coefficients are estimated directly from the paired observations, without coefficient sweeps or iterative optimization, and the resulting maps are fitted once and reused across subsequent samples.
At inference, the frozen maps act on online reference accelerations and correct cached residuals through lightweight elementwise operations, preserving the original cache schedule without additional full-network evaluations or online optimization.
A controlled DiT-XL/2 study (DDIM-20, $N=5$, 50K images per variant) further supports acceleration as the calibration target: under the same two-anchor update and deployment policy, acceleration supervision improves all five distributional metrics over residual and velocity supervision.

We evaluate \method{} on FLUX.1-dev and Qwen-Image using DrawBench with 20-step sampling at $N=3,4,5$.
Across these intervals, it improves PSNR, SSIM, and LPIPS over TaylorSeer and HiCache on both models while reducing denoising latency.
On FLUX.1-dev at $N=5$, ImageReward increases from $0.775$ to $0.899$, with a $2.86\times$ speedup over full computation versus TaylorSeer's $2.69\times$ under the same cache schedule.
Figure~\ref{fig:intro_qualitative} illustrates closer preservation of Native appearance in three matched-prompt examples.
On Qwen-Image, CLIP and ImageReward also improve across all three intervals.
DiT-XL/2 experiments on ImageNet-256 with DDIM-20 further show lower FID and sFID than TaylorSeer at $N=3,4,5$.
Our contributions are as follows:
\begin{itemize}[leftmargin=1.4em,itemsep=2pt,topsep=3pt]
  \item We formulate paired cache-to-full acceleration supervision by matching observations from independently evolving full-compute and naive-cache trajectories.
  A discrete error identity motivates the target; a controlled DiT study compares supervision choices within a shared two-anchor predictor.

  \item We design \method{}, which estimates grouped scalar corrections by closed-form centered least squares from one trajectory pair per deployment configuration.
  The frozen maps preserve the cache schedule without additional full-network evaluations or online optimization.

  \item We evaluate \method{} on FLUX.1-dev, Qwen-Image, and DiT-XL/2.
  Compared with TaylorSeer and HiCache, \method{} improves fidelity to full-compute outputs and reduces denoising latency on both text-to-image models, while achieving lower FID and sFID on DiT-XL/2.
\end{itemize}

%% file: sections/02_related_work.tex
\section{Related Work}

\paragraph{Cache policies.}
Cache policies determine when and where to skip network computation.
DeepCache~\citep{ma2023deepcache} reuses high-level U-Net features across denoising steps while updating low-level features, exploiting temporal redundancy.
$\Delta$-DiT~\citep{chen2024deltadit} combines residual caching with stage-dependent block selection, skipping rear blocks during early sampling and front blocks during later sampling.
ToCa~\citep{zou2025toca} selects tokens for reuse according to temporal redundancy and error propagation, with different caching ratios across layer types and depths.
TeaCache~\citep{liu2025teacache} estimates output changes from timestep-modulated input differences, using rescaled estimates to decide when cached outputs should be refreshed.
SeaCache~\citep{chung2026seacache} filters input features to suppress noise and retain content-relevant components, producing redundancy estimates that account for spectral evolution during denoising.
ERTACache~\citep{peng2026ertacache} combines offline cache-policy selection and timestep adjustment with approximate closed-form rectification of cached model-output errors.

\paragraph{Feature prediction.}
FORA~\citep{selvaraju2024fora} directly reuses cached attention and MLP outputs, yielding a piecewise-constant approximation between full-compute steps.
GOC~\citep{qiu2025goc} corrects cached attention and MLP features using recent temporal differences with a prescribed extrapolation scale, while feature statistics and timestep positions determine when to apply correction.
TaylorSeer~\citep{liu2025taylorseer} estimates higher-order feature derivatives from historical observations using finite differences and extrapolates future features through Taylor expansions.
HiCache~\citep{feng2026hicache} predicts features with Hermite polynomials and introduces dual scaling of the prediction horizon and higher-order terms for numerical stability.
FoCa~\citep{zheng2025foca} treats feature evolution as an ODE, combining multistep forecasting with a lightweight corrector anchored to the latest fully computed feature.
TC-Pad\'e~\citep{cui2026tcpade} models residual evolution through rational functions, adapting coefficients to historical changes and prediction strategies to different denoising stages.

Offline learning fits how cached features should be combined or transformed using full-compute supervision.
LearniBridge~\citep{huang2026learnibridge} trains low-rank adapters in the final Transformer block to reconstruct skipped-step representations from cached inputs under full-compute supervision, bypassing earlier blocks during inference.
$L^{2}P$~\citep{shen2026l2p} learns timestep-specific linear combinations of historical features from full-compute trajectories, replacing formula-derived coefficients with weights optimized for feature reconstruction.
\method{} instead calibrates feature acceleration using aligned observations from independently evolving full-compute and naive-cache trajectories, with scalar maps fitted in closed form from one trajectory pair.

%% file: sections/03_method.tex
\section{Method}
\label{sec:method}
\begin{figure}[t]
\centering
\includegraphics[width=\linewidth]{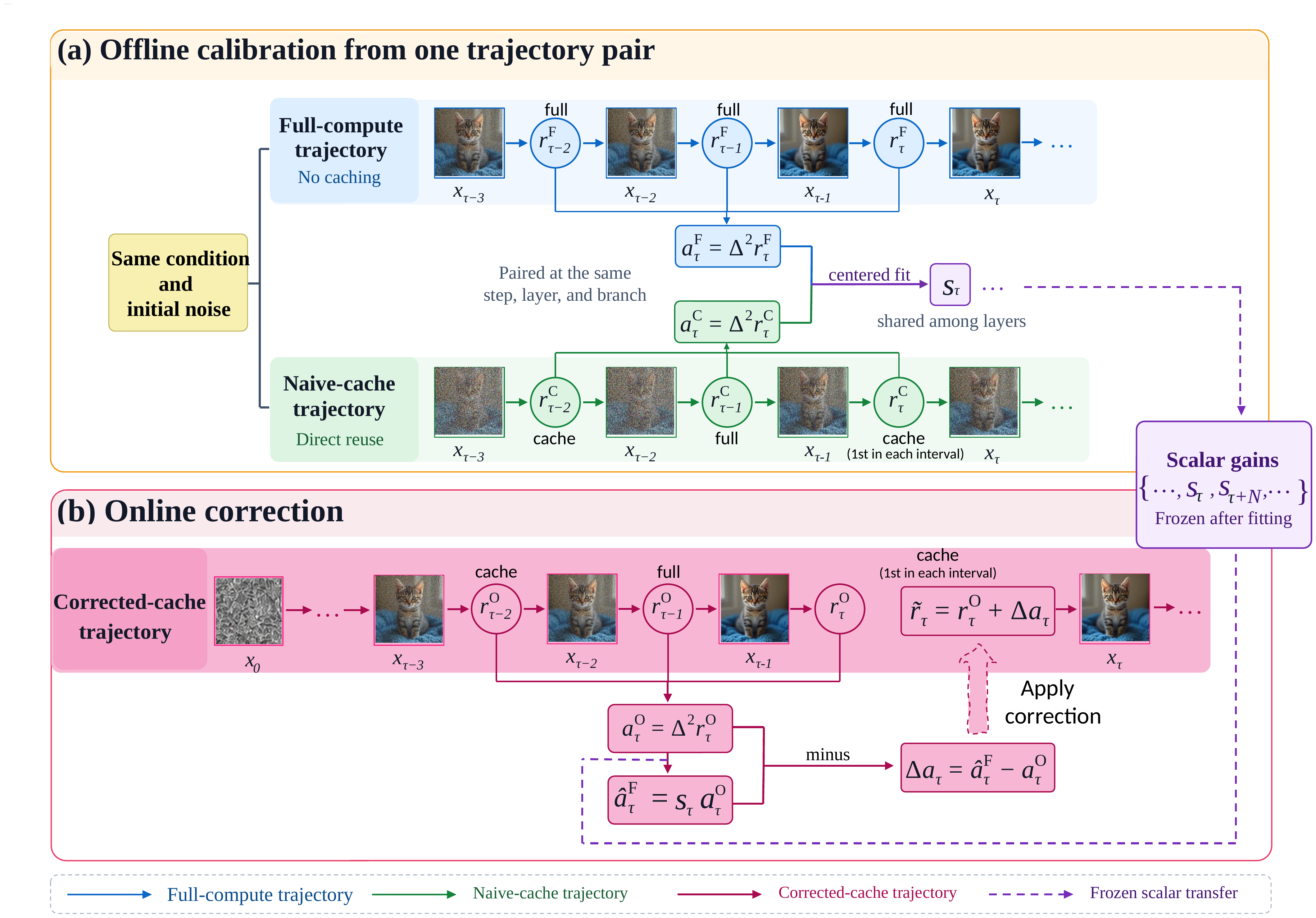}
\caption{Overview of \method{}. (a) One full-compute/naive-cache trajectory pair with the same condition and initial noise provides acceleration observations for centered scalar fitting at the first cache step of each interval. (b) On the online corrected-cache rollout, the frozen scalars estimate full-compute acceleration; the difference from the online reference acceleration corrects the current cached residual without additional full-network evaluations.}
\label{fig:overview}
\end{figure}

For a fixed model and cache schedule, \method{} calibrates discrepancies between cached and full-compute feature acceleration.
As shown in Figure~\ref{fig:overview}, one calibration sample supplies a full-compute/naive-cache trajectory pair for fitting grouped scalar maps offline; the frozen maps then correct residuals along an online corrected-cache rollout.
The calibration and correction target the first cache step of each interval.

\subsection{Feature acceleration}
\label{sec:feature-acceleration}

Consider a diffusion or flow-matching sampler with $T$ sampling steps, indexed by $\tau=0,\ldots,T-1$ in execution order from noise to data.
A \emph{branch} refers to a residual-producing submodule within a Transformer layer, such as an attention or MLP submodule.
Let $\br_{\tau,\ell,b}^{\full}$ and $\br_{\tau,\ell,b}^{\cache}$ denote the residual features of branch $b$ in layer $\ell$ at sampling step $\tau$ along the full-compute and naive-cache trajectories, respectively.
Let $\mathcal{S}$ denote a prescribed cache schedule that specifies which
branch evaluations are skipped, and let $N$ denote its nominal cache interval.
The full-compute trajectory evaluates every branch at every step.
The naive-cache trajectory follows $\mathcal{S}$, evaluating branches normally at full-compute steps and directly reusing their latest full-compute residuals at cache steps.
For either trajectory, suppressing the layer and branch indices, define the
first- and second-order finite differences for $\tau\geq 1$ and $\tau\geq 2$,
respectively, as
\begin{equation}
\bv_{\tau}=\br_{\tau}-\br_{\tau-1},\qquad
\ba_{\tau}=\Delta^2\br_{\tau}
=\br_{\tau}-2\br_{\tau-1}+\br_{\tau-2}.
\label{eq:acceleration}
\end{equation}
We refer to $\ba_{\tau}$ as feature acceleration. This term denotes a second finite difference of features and does not imply a continuous-time derivative.
The superscripts $\full$ and $\cache$ distinguish full-compute and naive-cache quantities: $\ba_{\tau,\ell,b}^{\full}$ and $\ba_{\tau,\ell,b}^{\cache}$ are computed from the residual sequences of their respective trajectories using Equation~(\ref{eq:acceleration}).

\paragraph{Acceleration calibration.}
We address the discrepancy between naive-cache and full-compute feature accelerations introduced by residual reuse.
We fit grouped scalar maps from naive-cache accelerations to their full-compute counterparts using paired calibration trajectories.
The fitted maps are then used to correct cached residuals at the first cache step of each interval.
We compare acceleration supervision with residual and velocity supervision in Table~\ref{tab:ablation-correction-target}; the fitting targets and shared deployment protocol are specified in Appendix~\ref{app:mechanism-details}.

\subsection{Offline calibration}
\label{sec:offline-calibration}

We fix one conditioning signal (a text prompt or class label), initial noise, guidance, solver, and timestep schedule, then record a full-compute trajectory and a naive-cache trajectory under the prescribed cache schedule.
The trajectories evolve independently; at the first cache step of each interval, accelerations computed from their respective residual sequences are paired for the same sampling step $\tau$, layer $\ell$, and branch $b$, as in Figure~\ref{fig:overview}(a).
This pairing matches the step, layer, and branch indices across trajectories, without requiring their intermediate latent states to coincide: each observation is extracted from the state reached by its own rollout.

\paragraph{Centered scalar fitting.}
We group observations by sampling step and computational branch, $g=G(\tau,b)$, and share each scalar across layers and tokens in its group.
Let $\mathcal{I}_g$ index the paired channel vectors $(\ba_j^{\cache},\ba_j^{\full})$ at corresponding layers and token positions.
Each layer and token position contributes a paired vector, so a single trajectory pair supplies many observations for estimating one scalar.
This sharing retains dependence on the sampling step and computational branch while restricting the fitted map to a common scale within each group.
Define the group means as $\bar{\ba}_g^{\star}=|\mathcal{I}_g|^{-1}\sum_{j\in\mathcal{I}_g}\ba_j^{\star}$ for $\star\in\{\full,\cache\}$.
We estimate a shared scale from centered observations:
\begin{equation}
s_g=\operatorname*{arg\,min}_{s\in\mathbb{R}}
\sum_{j\in\mathcal{I}_g}
\left\|(\ba_j^{\full}-\bar{\ba}_g^{\full})
-s(\ba_j^{\cache}-\bar{\ba}_g^{\cache})\right\|_2^2.
\label{eq:ols}
\end{equation}
For nonzero centered input energy, the closed-form solution is
\begin{equation}
s_g=
\frac{\displaystyle\sum_{j\in\mathcal{I}_g}
\left\langle\ba_j^{\cache}-\bar{\ba}_g^{\cache},
\ba_j^{\full}-\bar{\ba}_g^{\full}\right\rangle}
{\displaystyle\sum_{j\in\mathcal{I}_g}
\left\|\ba_j^{\cache}-\bar{\ba}_g^{\cache}\right\|_2^2}.
\label{eq:closedform}
\end{equation}
The numerator measures alignment between centered accelerations, normalized by the cached energy in the denominator.
Thus, the shared scale is estimated from within-group co-variation rather than group means.
The deployed map is scalar multiplication by $s_g$; the means are used only during fitting.
We write $s_g=s_{\tau,b}$ for $g=G(\tau,b)$ and suppress the branch index as $s_\tau$ in Figure~\ref{fig:overview} and the online equations below.
The fitted scalars remain frozen for all subsequent samples under the same deployment configuration.

\subsection{Online correction}
\label{sec:online-correction}

Let $\br_{\tau,\ell,b}^{\mathrm{O}}$ denote the online reference residual supplied to the calibration operator, equal to the current pre-correction residual at a correction step.
The preceding $\br_{\tau-1,\ell,b}^{\mathrm{O}}$ and $\br_{\tau-2,\ell,b}^{\mathrm{O}}$ belong to this reference sequence for the same layer and branch, while $\widetilde{\br}_{\tau,\ell,b}$ is the corrected residual feature.
The superscript $\mathrm{O}$ distinguishes online references from the offline full-compute ($\full$) and naive-cache ($\cache$) trajectories.
Suppressing layer and branch indices, at an eligible first cache step with two preceding references, we compute $\ba_\tau^{\mathrm{O}}=\Delta^2\br_\tau^{\mathrm{O}}$ using Equation~(\ref{eq:acceleration}).
Following Figure~\ref{fig:overview}(b), the frozen map gives a full-compute acceleration estimate, whose discrepancy from the reference acceleration corrects the current residual:
\begin{equation}
\begin{aligned}
\widehat{\ba}_\tau^{\full}
&=s_\tau\ba_\tau^{\mathrm{O}},
\qquad \Delta\ba_\tau=\widehat{\ba}_\tau^{\full}-\ba_\tau^{\mathrm{O}},\\
\widetilde{\br}_\tau
&=\br_\tau^{\mathrm{O}}+\Delta\ba_\tau
=\br_\tau^{\mathrm{O}}+(s_\tau-1)\ba_\tau^{\mathrm{O}}.
\end{aligned}
\label{eq:update}
\end{equation}
Here $\Delta\ba_\tau$ is the correction discrepancy and $\widehat{\ba}_\tau^{\full}$ is a mapped estimate, not an additional full-compute evaluation.
With the preceding references fixed, adding $\Delta\ba_\tau$ changes the current residual's second difference by the same amount.
This local relation is defined relative to the reference sequence; the residual sequence actually used by the network also reflects the deployment policy.
The correction remains sample-dependent through $\ba_\tau^{\mathrm{O}}$, while $s_\tau=1$ leaves the current residual unchanged.

The correction operator itself uses a three-point finite difference and elementwise scaling and addition, costing $O(d)$ for a $d$-dimensional residual token.
The coefficient storage is $O(|\mathcal{G}|)$ for step--branch groups $\mathcal{G}$, with one scalar per group; runtime feature buffers are separate.
The operator preserves the full/cache schedule and requires neither additional full-network evaluations nor online optimization.
Implementation details and deployment-policy analyses are provided in Section~\ref{sec:experiments} and Appendix~\ref{app:implementation}.

%% file: sections/04_experiments.tex
\section{Experiments}
\label{sec:experiments}

\subsection{Experimental Setup}
\label{sec:experimental-setup}

\paragraph{Models and benchmarks.}
We evaluate 20-step sampling at cache intervals $N\in\{3,4,5\}$ on three diffusion Transformers.
FLUX.1-dev~\citep{blackforestlabs2024flux} uses shifted flow-matching steps and guidance 3.5; Qwen-Image~\citep{wu2025qwenimage} uses true classifier-free guidance (CFG) 4.0.
Both text-to-image models are evaluated on DrawBench-200~\citep{saharia2022imagen} at $1024^2$ resolution.
DiT-XL/2~\citep{peebles2023dit} is evaluated on ImageNet-256 using DDIM-20~\citep{song2021ddim}, CFG 1.5, and 50K generated images per configuration.
Each model--interval configuration uses one calibration condition (a COCO~\citep{lin2014coco} caption disjoint from evaluation prompts, or a class label) and fixed initial noise to generate a full-compute/naive-cache trajectory pair. The centered estimator in Equation~(\ref{eq:closedform}) fits one scalar per correction step and branch, reused across the entire evaluation set.

The model-named baseline rows use full computation at all 20 sampling steps.
Calibration and deployment details are given in Appendix~\ref{app:deployment}.
\paragraph{Baselines.}
We compare \method{} with FORA direct reuse~\citep{selvaraju2024fora} and two history-based forecasting baselines, TaylorSeer~\citep{liu2025taylorseer} and HiCache~\citep{feng2026hicache}, both at order $O=2$.
HiCache uses the authors' official scaled-Hermite predictor (commit \texttt{4d0839f}) with scale $0.6$.
\paragraph{Unified deployment policy.}
Across all three models, TaylorSeer, HiCache, and \method{} share the same full/cache schedule: warmup-2 at $N=3$ and warmup-3 at $N=4,5$, followed by full evaluations every $N$ steps from the last warmup step. FORA instead refreshes every $N$ steps from the initial full-compute step, without extra warmup.
\method{} applies Equation~(\ref{eq:update}) only at the first cache step of each interval, writes the corrected residual into the reuse buffer, and reuses it until the next full-compute step.
A separate predictor history records the latest full-compute anchor at every cache step, supplying the references $\br^{\mathrm{O}}$ in Section~\ref{sec:online-correction} and isolating direct feedback from corrected reuse.
Full-compute steps refresh both the anchor and the reuse buffer on the deployed rollout.
At $N=4$, the final step (logical step $0$) is a first cache step; we skip its correction to avoid a large terminal perturbation that could degrade image quality. Appendix~\ref{app:deployment} lists the exact step positions.
\paragraph{Metrics and efficiency.}
PSNR, SSIM, and LPIPS~\citep{zhang2018lpips} compare each final image with the full-compute output generated from the same prompt and seed, measuring \emph{full-compute fidelity}.
CLIP~\citep{radford2021clip} and ImageReward (IR)~\citep{xu2023imagereward} assess semantic alignment and learned human-preference scores, respectively. 
For DiT-XL/2 on ImageNet-256, we report FID~\citep{heusel2017fid}, sFID, IS, precision, and recall.
We measure per-image denoising latency on one NVIDIA H800 GPU at batch size one, averaging 30 CUDA-event measurements after ten benchmark warmup runs.
Model weights and calibration parameters are GPU-resident before timing, which excludes text encoding, VAE decoding, pack preparation, and disk I/O.
FLOPs cover the same 20-step denoising-only region as latency. We use \texttt{torch.profiler} for profiler-visible tensor operations, including linear projections, and analytically add fused attention matrix multiplications, custom Triton correction kernels, and TaylorSeer/HiCache predictor arithmetic omitted by the profiler. Conditioning, VAE decoding, device-to-host transfer, and disk I/O are excluded for all three models. The accounting scope is specified in Appendix~\ref{app:efficiency}.

\subsection{FLUX DrawBench}
\label{sec:flux-drawbench}

Table~\ref{tab:drawbench} compares the deployed implementations of \method{}, FORA, TaylorSeer, and HiCache.
Across $N=3,4,5$, \method{} achieves the best PSNR, SSIM, LPIPS, and IR among the evaluated caching methods.
It also obtains the highest CLIP score at $N=4,5$.

At $N=5$, \method{} increases PSNR from $17.2693$ for TaylorSeer and $17.9516$ for HiCache to $19.1014$ and increases IR from $0.77485$ and $0.84138$, respectively, to $0.89891$.
Its denoising speedup reaches $2.86\times$, compared with $2.69\times$ for TaylorSeer and $2.43\times$ for HiCache.
Across all three intervals, \method{} is faster than TaylorSeer; its latency is comparable to HiCache at $N=3$ and lower at $N=4,5$.
The reported FLOPs include the additional single-stream projections in the FLUX implementation.
Table~\ref{tab:ablation-uncorrected-reuse} further compares \method{} with uncorrected reuse at matched full/cache positions.

\begin{table}[t]
\centering
\caption{FLUX.1-dev on DrawBench-200 at $1024^2$ with 20 denoising steps.}
\label{tab:drawbench}
\scriptsize
\setlength{\tabcolsep}{2.5pt}
\renewcommand{\arraystretch}{1.10}
\begin{tabular*}{\linewidth}{@{\extracolsep{\fill}}l c c c c c c c@{}}
\toprule
\multirow{2}{*}{Method} & \multicolumn{2}{c}{Efficiency} & \multicolumn{2}{c}{Alignment / preference} & \multicolumn{3}{c}{Full-compute fidelity} \\
\cmidrule(lr){2-3}\cmidrule(lr){4-5}\cmidrule(lr){6-8}
& Latency (s)$\downarrow$ & FLOPs (T)$\downarrow$ & CLIP$\uparrow$ & IR$\uparrow$ & PSNR$\uparrow$ & SSIM$\uparrow$ & LPIPS$\downarrow$ \\
\midrule
FLUX.1-dev (20 steps) & 7.0495 (1.00$\times$) & 1487.836 (1.00$\times$) & 31.3273 & 0.95212 & $\infty$ & 1 & 0 \\
\midrule
FORA ($N=3$) & \best{2.7235} (2.59$\times$) & \best{521.051} (2.86$\times$) & 31.2644 & 0.7261 & 15.9921 & 0.64406 & 0.47070 \\
TaylorSeer ($N=3,O=2$) & 3.2887 (2.14$\times$) & 595.719 (2.50$\times$) & \best{31.5913} & 0.93172 & 16.7465 & 0.66782 & 0.41637 \\
HiCache ($N=3,O=2$) & 3.2052 (2.20$\times$) & 595.539 (2.50$\times$) & 31.5534 & 0.93172 & 17.0422 & 0.66129 & 0.41414 \\
\method{} ($N=3$) & 3.2094 (2.20$\times$) & 621.914 (2.39$\times$) & 31.5376 & \best{0.95549} & \best{18.6202} & \best{0.72185} & \best{0.34093} \\
\midrule
FORA ($N=4$) & \best{2.0177} (3.49$\times$) & \best{372.315} (4.00$\times$) & 31.0429 & 0.4337 & 15.3897 & 0.59587 & 0.54240 \\
TaylorSeer ($N=4,O=2$) & 2.9441 (2.39$\times$) & 521.346 (2.85$\times$) & 31.5866 & 0.93917 & 18.5928 & 0.71370 & 0.34726 \\
HiCache ($N=4,O=2$) & 3.1877 (2.21$\times$) & 521.155 (2.85$\times$) & 31.5430 & 0.94229 & 19.1413 & 0.71836 & 0.33004 \\
\method{} ($N=4$) & 2.8173 (2.50$\times$) & 544.223 (2.73$\times$) & \best{31.6116} & \best{0.96672} & \best{19.8328} & \best{0.74463} & \best{0.28787} \\
\midrule
FORA ($N=5$) & \best{1.6694} (4.22$\times$) & \best{297.947} (4.99$\times$) & 30.3692 & 0.0469 & 15.0238 & 0.55455 & 0.60742 \\
TaylorSeer ($N=5,O=2$) & 2.6177 (2.69$\times$) & 446.987 (3.33$\times$) & 31.8857 & 0.77485 & 17.2693 & 0.59087 & 0.56288 \\
HiCache ($N=5,O=2$) & 2.8977 (2.43$\times$) & 446.783 (3.33$\times$) & 31.6713 & 0.84138 & 17.9516 & 0.64256 & 0.44769 \\
\method{} ($N=5$) & 2.4644 (2.86$\times$) & 466.550 (3.19$\times$) & \best{31.9461} & \best{0.89891} & \best{19.1014} & \best{0.67348} & \best{0.42925} \\
\bottomrule
\end{tabular*}
\end{table}

\subsection{Qwen-Image DrawBench}
\label{sec:qwen-drawbench}

On Qwen-Image, \method{} improves all five reported alignment, preference, and fidelity metrics over FORA, TaylorSeer, and HiCache at $N=3,4,5$ (Table~\ref{tab:qwen-drawbench}).
At $N=5$, LPIPS decreases from $0.50306$ for TaylorSeer and $0.47577$ for HiCache to $0.30412$, corresponding to reductions of $39.5\%$ and $36.1\%$.
ImageReward increases from $0.45596$ and $0.52504$, respectively, to $1.00084$.
\method{} also has lower denoising latency than both TaylorSeer and HiCache at every interval, reaching $2.27\times$ speedup at $N=5$ versus $2.01\times$ and $1.80\times$, respectively.

\begin{table}[t]
\centering
\caption{Qwen-Image on DrawBench-200 at $1024^2$ with 20 denoising steps.}
\label{tab:qwen-drawbench}
\scriptsize
\setlength{\tabcolsep}{2.5pt}
\renewcommand{\arraystretch}{1.10}
\begin{tabular*}{\linewidth}{@{\extracolsep{\fill}}l c c c c c c c@{}}
\toprule
\multirow{2}{*}{Method} & \multicolumn{2}{c}{Efficiency} & \multicolumn{2}{c}{Alignment / preference} & \multicolumn{3}{c}{Full-compute fidelity} \\
\cmidrule(lr){2-3}\cmidrule(lr){4-5}\cmidrule(lr){6-8}
& Latency (s)$\downarrow$ & FLOPs (T)$\downarrow$ & CLIP$\uparrow$ & IR$\uparrow$ & PSNR$\uparrow$ & SSIM$\uparrow$ & LPIPS$\downarrow$ \\
\midrule
Qwen-Image (20 steps) & 12.7757 (1.00$\times$) & 2728.591 (1.00$\times$) & 32.8989 & 1.17516 & $\infty$ & 1 & 0 \\
\midrule
FORA ($N=3$) & \best{6.1979} (2.06$\times$) & \best{955.599} (2.86$\times$) & 31.0854 & 0.3168 & 13.9155 & 0.54600 & 0.53277 \\
TaylorSeer ($N=3,O=2$) & 7.4616 (1.71$\times$) & 1092.537 (2.50$\times$) & 32.1725 & 0.98366 & 13.4594 & 0.56454 & 0.49408 \\
HiCache ($N=3,O=2$) & 8.0308 (1.59$\times$) & 1092.170 (2.50$\times$) & 32.0616 & 0.98763 & 13.3377 & 0.56193 & 0.48487 \\
\method{} ($N=3$) & 6.7595 (1.89$\times$) & 1091.973 (2.50$\times$) & \best{32.9019} & \best{1.10332} & \best{18.1881} & \best{0.70563} & \best{0.31579} \\
\midrule
FORA ($N=4$) & \best{5.1507} (2.48$\times$) & \best{682.814} (4.00$\times$) & 30.2783 & -0.2850 & 13.4366 & 0.51479 & 0.58241 \\
TaylorSeer ($N=4,O=2$) & 7.0275 (1.82$\times$) & 956.136 (2.85$\times$) & 31.5651 & 0.75211 & 15.3328 & 0.64752 & 0.41676 \\
HiCache ($N=4,O=2$) & 7.6527 (1.67$\times$) & 955.747 (2.85$\times$) & 31.7830 & 0.79409 & 15.7373 & 0.65402 & 0.38992 \\
\method{} ($N=4$) & 6.1706 (2.07$\times$) & 955.551 (2.86$\times$) & \best{32.6364} & \best{1.11191} & \best{20.4604} & \best{0.78839} & \best{0.20340} \\
\midrule
FORA ($N=5$) & \best{4.6055} (2.77$\times$) & \best{546.422} (4.99$\times$) & 28.6520 & -0.8597 & 13.1403 & 0.49811 & 0.61961 \\
TaylorSeer ($N=5,O=2$) & 6.3679 (2.01$\times$) & 819.762 (3.33$\times$) & 30.6241 & 0.45596 & 13.7709 & 0.59085 & 0.50306 \\
HiCache ($N=5,O=2$) & 7.0980 (1.80$\times$) & 819.351 (3.33$\times$) & 30.9754 & 0.52504 & 14.0766 & 0.59470 & 0.47577 \\
\method{} ($N=5$) & 5.6295 (2.27$\times$) & 819.158 (3.33$\times$) & \best{32.4704} & \best{1.00084} & \best{19.2460} & \best{0.71478} & \best{0.30412} \\
\bottomrule
\end{tabular*}
\end{table}

\subsection{DiT-XL/2 on ImageNet-256}
\label{sec:dit}

Table~\ref{tab:dit} evaluates class-conditional generation under the unified deployment policy in Section~\ref{sec:experimental-setup}.
Each \method{} setting uses one calibration trajectory pair.
Across $N=3,4,5$, \method{} outperforms FORA, TaylorSeer, and HiCache on all five reported distributional metrics.
At $N=5$, FID decreases from $16.3916$ for TaylorSeer and $24.1568$ for HiCache to $10.3691$, while sFID decreases from $17.5930$ and $26.8663$, respectively, to $6.4061$.

\method{} is also faster than both TaylorSeer and HiCache at all three intervals, reaching $1.81\times$ speedup at $N=5$ compared with $1.64\times$ and $1.21\times$, respectively.
FORA offers higher speedup but has substantially worse FID and sFID.

\begin{table}[t]
\centering
\caption{DiT-XL/2 on ImageNet-256 with DDIM-20 and 50K-image quality evaluation.}
\label{tab:dit}
\label{tab:dit-extended-20steps}
\scriptsize
\setlength{\tabcolsep}{2.5pt}
\renewcommand{\arraystretch}{1.10}
\begin{tabular*}{\linewidth}{@{\extracolsep{\fill}}l c c c c c c c@{}}
\toprule
\multirow{2}{*}{Method} & \multicolumn{2}{c}{Efficiency} & \multicolumn{5}{c}{Distributional evaluation} \\
\cmidrule(lr){2-3}\cmidrule(lr){4-8}
& Latency (s)$\downarrow$ & FLOPs (T)$\downarrow$ & FID$\downarrow$ & sFID$\downarrow$ & IS$\uparrow$ & Precision$\uparrow$ & Recall$\uparrow$ \\
\midrule
DiT-XL/2 (DDIM-20) & 0.2449 (1.00$\times$) & 9.1535 (1.00$\times$) & 3.4931 & 4.9673 & 223.1583 & 0.78447 & 0.57610 \\
\midrule
FORA ($N=3$) & \best{0.1212} (2.02$\times$) & \best{3.2167} (2.85$\times$) & 14.2871 & 17.0764 & 144.4846 & 0.65272 & 0.47860 \\
TaylorSeer ($N=3,O=2$) & 0.1761 (1.39$\times$) & 3.6804 (2.49$\times$) & 5.3006 & 7.2121 & 196.4148 & 0.75718 & 0.55490 \\
HiCache ($N=3,O=2$) & 0.2220 (1.10$\times$) & 3.6759 (2.49$\times$) & 7.8976 & 10.8324 & 176.9213 & 0.72976 & 0.51910 \\
\method{} ($N=3$) & 0.1662 (1.47$\times$) & 3.6766 (2.49$\times$) & \best{4.1941} & \best{6.1416} & \best{207.6349} & \best{0.75742} & \best{0.59240} \\
\midrule
FORA ($N=4$) & \best{0.1049} (2.33$\times$) & \best{2.3034} (3.97$\times$) & 30.0836 & 36.5201 & 85.7020 & 0.51546 & 0.41770 \\
TaylorSeer ($N=4,O=2$) & 0.1622 (1.51$\times$) & 3.2237 (2.84$\times$) & 8.4713 & 9.6293 & 170.9763 & 0.71073 & 0.53280 \\
HiCache ($N=4,O=2$) & 0.2098 (1.17$\times$) & 3.2189 (2.84$\times$) & 12.6385 & 14.8089 & 148.1833 & 0.66596 & 0.50220 \\
\method{} ($N=4$) & 0.1453 (1.69$\times$) & 3.2189 (2.84$\times$) & \best{6.7296} & \best{6.0563} & \best{179.9203} & \best{0.71430} & \best{0.58140} \\
\midrule
FORA ($N=5$) & \best{0.0969} (2.53$\times$) & \best{1.8467} (4.96$\times$) & 48.1062 & 52.2517 & 51.5605 & 0.39456 & 0.39680 \\
TaylorSeer ($N=5,O=2$) & 0.1497 (1.64$\times$) & 2.7673 (3.31$\times$) & 16.3916 & 17.5930 & 127.8015 & 0.62528 & 0.48940 \\
HiCache ($N=5,O=2$) & 0.2028 (1.21$\times$) & 2.7621 (3.31$\times$) & 24.1568 & 26.8663 & 100.4643 & 0.55762 & 0.45810 \\
\method{} ($N=5$) & 0.1354 (1.81$\times$) & 2.7622 (3.31$\times$) & \best{10.3691} & \best{6.4061} & \best{151.7564} & \best{0.66828} & \best{0.57380} \\
\bottomrule
\end{tabular*}
\end{table}

%% file: sections/05_ablation.tex
\section{Ablation Studies}
\label{sec:mechanism}

\paragraph{Correction target.}
We compare residual, velocity, and acceleration supervision on DiT-XL/2 (ImageNet-256, DDIM-20, $N=5$, warmup-3).
All variants use the same two-anchor update, step-by-branch centered fitting, and calibration trajectory pair; only the coefficient supervision changes.
They share first-cache correction, corrected writeback, and isolated full-anchor history.
Table~\ref{tab:ablation-correction-target} reports 50K-image distributional metrics.
Acceleration supervision achieves FID $10.36915$, versus $41.80262$ for residual and $37.27242$ for velocity supervision, and leads all five metrics in this controlled DiT setting.
Appendix~\ref{app:mechanism-details} specifies the shared predictor and fitting targets.
\begin{table}[t]
\centering
\caption{Correction-target ablation on DiT-XL/2, ImageNet-256, DDIM-20, $N=5$, warmup-3. All variants share a two-anchor predictor, one calibration trajectory pair, centered scalar fitting, first-cache correction, corrected writeback, isolated history, and 50K-image evaluation. Only the target used to fit the scalar coefficients differs: residual, velocity, or acceleration.}
\label{tab:ablation-correction-target}
\label{tab:mechanism-target}
\small
\setlength{\tabcolsep}{8pt}
\renewcommand{\arraystretch}{1.18}
\begin{tabular}{@{}l r r r r r@{}}
\toprule
Supervision & FID$\downarrow$ & sFID$\downarrow$ & IS$\uparrow$ & Precision$\uparrow$ & Recall$\uparrow$ \\
\midrule

Residual supervision & 41.80262 & 45.89258 & 64.36098 & 0.42992 & 0.42250 \\
Velocity ($v$) & 37.27242 & 40.45567 & 72.52494 & 0.45670 & 0.43760 \\
Acceleration ($a$) & \best{10.36915} & \best{6.40612} & \best{151.75644} & \best{0.66828} & \best{0.57380} \\
\bottomrule
\end{tabular}
\end{table}

\begin{table}[t]
\centering
\caption{Matched-schedule comparison with uncorrected reuse at $N=5$, warmup-3, and 20 steps. FLUX.1-dev and Qwen-Image use DrawBench-200 at $1024^2$; DiT-XL/2 uses ImageNet-256, DDIM-20, and 50K images per row. Full/cache positions are matched; denoising latency and FLOPs reflect each implementation.}
\label{tab:ablation-uncorrected-reuse}
\scriptsize
\setlength{\tabcolsep}{4pt}
\resizebox{\textwidth}{!}{%
\begin{tabular}{l c c c c c c c}
\toprule
\textbf{FLUX.1-dev} & Latency (s)$\downarrow$ & FLOPs (T)$\downarrow$ & CLIP$\uparrow$ & IR$\uparrow$ & PSNR$\uparrow$ & SSIM$\uparrow$ & LPIPS$\downarrow$ \\
\midrule
Uncorrected reuse & \best{2.2425} & \best{446.683} & 31.5936 & 0.84588 & 18.7906 & 0.60969 & 0.45937 \\
\method{} & 2.4644 & 466.550 & \best{31.9461} & \best{0.89891} & \best{19.1014} & \best{0.67348} & \best{0.42925} \\
\midrule
\textbf{Qwen-Image} & Latency (s)$\downarrow$ & FLOPs (T)$\downarrow$ & CLIP$\uparrow$ & IR$\uparrow$ & PSNR$\uparrow$ & SSIM$\uparrow$ & LPIPS$\downarrow$ \\
\midrule
Uncorrected reuse & \best{5.4920} & \best{819.098} & \best{32.5109} & 0.91775 & 18.1857 & 0.65986 & 0.30701 \\
\method{} & 5.6295 & 819.158 & 32.4704 & \best{1.00084} & \best{19.2460} & \best{0.71478} & \best{0.30412} \\
\midrule
\textbf{DiT-XL/2} & Latency (s)$\downarrow$ & FLOPs (T)$\downarrow$ & FID$\downarrow$ & sFID$\downarrow$ & IS$\uparrow$ & Precision$\uparrow$ & Recall$\uparrow$ \\
\midrule
Uncorrected reuse & \best{0.1128} & \best{2.7601} & 45.7141 & 48.7345 & 57.8806 & 0.40062 & 0.43130 \\
\method{} & 0.1354 & 2.7622 & \best{10.3691} & \best{6.4061} & \best{151.7564} & \best{0.66828} & \best{0.57380} \\
\bottomrule
\end{tabular}}
\end{table}

\Needspace{5\baselineskip}
\paragraph{Matched-schedule comparison with uncorrected reuse.}
Table~\ref{tab:ablation-uncorrected-reuse} matches the six full-compute steps $\{0,1,2,7,12,17\}$ at $N=5$, warmup-3.
Unlike standard FORA, this reuse control includes the same sampling warmup; the \method{} rows repeat the complete main-table implementation.
On FLUX and DiT, all five quality metrics improve.
FLUX LPIPS decreases by $6.6\%$ with $9.9\%$ higher latency; DiT FID decreases by $77.3\%$ with $20.0\%$ higher latency.
On Qwen, SSIM rises from $0.65986$ to $0.71478$ and IR from $0.91775$ to $1.00084$ with $2.5\%$ higher latency, while CLIP decreases slightly.
The incremental quality--latency trade-off is therefore model- and metric-dependent.
Matching sampling positions does not impose identical operator-level computation: FLUX FLOPs also increase by $4.45\%$ due to the additional projections used to process text and image features separately in our single-stream implementation.
This comparison measures the deployed method against matched-schedule reuse, rather than isolating the fitted coefficient on an identical execution path.

\paragraph{Writeback and predictor-history isolation.}
Table~\ref{tab:dit-writeback-history} compares three application variants on DiT-XL/2 with DDIM-20, $N=5$, warmup-3, and CFG 1.5.
All variants use the same one-trajectory scalar pack, the correction operator in Equation~(\ref{eq:update}), and the same full/cache/correction positions, with 50K generated images per variant.
Correction without writeback applies the scalar correction at eligible steps but retains the full anchor in the reuse buffer and uses full-anchor predictor history.
Ordinary writeback reuses corrected residuals and records predictor history from that evolving reuse stream.
\method{} additionally isolates predictor history by recording the latest full-compute anchor at every cache step.

\begin{table}[!t]
\centering
\caption{Application-policy ablation on DiT-XL/2, ImageNet-256, DDIM-20, $N=5$. All rows use the same scalar calibration pack, warmup-3 schedule, and 50K-image evaluation protocol.}
\label{tab:dit-writeback-history}
\scriptsize
\setlength{\tabcolsep}{4pt}
\renewcommand{\arraystretch}{1.10}
\resizebox{\linewidth}{!}{%
\begin{tabular}{@{}l r r r r r@{}}
\toprule
Application policy & FID$\downarrow$ & sFID$\downarrow$ & IS$\uparrow$ & Precision$\uparrow$ & Recall$\uparrow$ \\
\midrule
Correction without writeback & 18.63283 & 13.06961 & 124.41991 & 0.59702 & 0.53530  \\
Ordinary writeback & 14.63578 & 9.01988 & 133.63596 & 0.62458 & 
\best{0.58280} \\
Writeback + isolated history (\method{}) & \best{10.36915} & \best{6.40612} & \best{151.75644} & \best{0.66828} & 0.57380\\
\bottomrule
\end{tabular}}
\end{table}

Writeback with isolated history reduces FID from $18.63283$ for correction without writeback and $14.63578$ for ordinary writeback to $10.36915$.
It also yields the lowest sFID and the highest IS and precision; ordinary writeback retains the highest recall ($0.58280$ versus $0.57380$).
These controlled results support corrected writeback with isolated history as the deployment policy.
Appendix~\ref{app:writeback-history} explains how the three policies update the reuse buffer and predictor history.

On FLUX.1-dev with 20 steps, $N=5$, and fixed calibration noise, ten random single-prompt calibrations yield small variation in evaluation-set aggregates; Appendix~\ref{app:additional-calibration-ablations} reports these statistics and calibration-set-size sensitivity.
Appendix~\ref{app:coefficient-choice} compares fixed and fitted coefficients on FLUX, with grouping further improving PSNR over a shared fit.

%% file: sections/05_conclusion.tex
\section{Conclusion}

We introduced \method{}, a lightweight offline method that improves feature caching in diffusion Transformers by calibrating cache-to-full feature acceleration.
Using a single pair of full-compute and naive-cache trajectories per configuration, it fits grouped scalar maps through closed-form centered least squares.
The frozen maps correct cached residuals while preserving the original cache schedule, without additional full-network evaluations or online optimization.
Under 20-step sampling, \method{} improves fidelity to full-compute outputs on FLUX.1-dev and Qwen-Image and reduces FID and sFID on DiT-XL/2 relative to TaylorSeer and HiCache.
Together, these results demonstrate the value of paired feature dynamics as supervision for lightweight correction.
This provides a practical way to improve the quality of cache-accelerated generation while retaining the computational savings of feature reuse, even when sampling already uses a short schedule.

%% file: sections/06_statements.tex
\section*{AI Use Statement}
Generative AI tools assisted with manuscript organization, language editing,
LaTeX formatting, literature retrieval and discovery (including identifying
relevant prior work), and cross-checking experimental records.
The authors have reviewed all AI-assisted content and checked the cited
sources and reported results against the original publications and
experimental records.
\subsection*{Ethics statement}
This work studies inference acceleration for generative models and involves neither human-subject experiments nor new collection of personal data. Lower inference cost may broaden access to generative models while also reducing barriers to misuse. Bias, safety, and copyright risks in generated outputs are inherited from the underlying models. We will retain the foundation-model licenses and known limitations in released code and documentation.
\subsection*{Reproducibility statement}
We prepared a minimal code package to support reproduction of our method, including paired-trajectory collection, centered scalar fitting, and calibrated sampling. Evaluation metrics are computed using their official implementations. The main paper and appendix specify the experimental settings, evaluation protocols, and implementation details.

%% file: sections/07_appendix_implementation.tex
\section{Implementation and Protocol Details}
\label{app:implementation}

\subsection{Calibration and Deployment Policies}
\label{app:deployment}
\paragraph{Numerical implementation of centered scalar fitting.}
The centered objective and its closed-form solution are given in Equations~(\ref{eq:ols}) and~(\ref{eq:closedform}).
The implementation accumulates sufficient statistics in double precision and guards numerically negligible centered energy with denominator floors or a zero-scale fallback.
Each group supplies one fitted scalar $s_g=s_{\tau,b}$ to the online map in Equation~(\ref{eq:update}); the fitting means are not required during sampling.

\paragraph{Calibration samples and parameter sharing.}
For the FLUX and Qwen-Image DrawBench experiments, each deployed pack is fitted using one COCO caption disjoint from the evaluation prompts. Each deployed DiT pack uses one calibration trajectory. The separate mechanism and sensitivity studies use the calibration counts stated in their own sections.
Single-trajectory calibration uses one conditioning/noise realization to generate a matched pair of full-compute and naive-cache trajectories. The conditioning (a prompt or DiT class label), initial noise, and sampler configuration are shared within the pair.
FLUX calibration uses seeds $314159+\text{global index}$; Qwen-Image evaluation uses $314159+\text{prompt index}$ consistently across methods.
FLUX uses step-by-branch packs for $N=3,4,5$.
Qwen-Image uses step-by-branch packs shared across CFG contexts, with separate conditional and unconditional cache states.
DiT uses step-by-branch maps for attention and MLP residuals, with each map shared across all 28 Transformer blocks.

\paragraph{Sampling warmup and cache schedules.}
Table~\ref{tab:warmup-protocol} specifies the sampling warmup, which is distinct from the ten execution warmup runs used for latency measurement.
The main-table FORA baseline uses no extra sampling warmup at any interval: its first denoising step is full, then it refreshes every $N$ steps. TaylorSeer, HiCache, and \method{} use warmup-2 at $N=3$ and warmup-3 at $N=4,5$ on all three models.
The Uncorrected reuse controls in the main-text Table~\ref{tab:ablation-uncorrected-reuse} use the same $N=5$/warmup-3 full/cache positions as \method{} on all three models.
The common warmup supplies the two preceding observations required by the first cache correction.

\begin{table}[t]
\centering
\caption{Sampling warmup: the initial full-compute steps before periodic caching. Zero denotes no additional warmup phase. The ordinary cache schedule still supplies full-compute evaluations.}
\label{tab:warmup-protocol}
\small
\setlength{\tabcolsep}{6pt}
\begin{tabular}{l l c c c}
\toprule
Model & Methods & $N=3$ & $N=4$ & $N=5$ \\
\midrule
FLUX / Qwen-Image & FORA & 0 & 0 & 0 \\
FLUX / Qwen-Image & TaylorSeer, HiCache, \method{} & 2 & 3 & 3 \\
DiT-XL/2 & FORA & 0 & 0 & 0 \\
DiT-XL/2 & TaylorSeer, HiCache, \method{} & 2 & 3 & 3 \\
\bottomrule
\end{tabular}
\end{table}

\paragraph{Terminal-step fallback.}
Logical steps count down from $19$ to $0$, whereas execution indices increase from $0$ to $19$.
At $N=4$, the final step is a first cache step; its correction is skipped to avoid a large terminal residual change that could degrade image quality, and the preceding full residual is reused directly.
This skip0 rule preserves all full/cache positions. At $N=3$ the final step is full, while at $N=5$ it only reuses the correction from the preceding cache step.

\subsubsection{Unified Interval-Wise Correction and Cache Writeback}
\label{app:dit-policy}

For all three models, \method{} applies Equation~(\ref{eq:update}) only at the first cache step in each interval, subject to the terminal-step fallback.
Suppressing layer and branch indices, let $\boldsymbol{f}_\tau$ denote the latest full-compute residual evaluated on the current deployed rollout.
The predictor history $H_{\mathrm{pred}}$ stores full-anchor reference residuals, while the deployment buffer $B_{\mathrm{reuse}}$ stores the residual to be reused by the network.
Full-compute steps refresh both the anchor and the reuse buffer and append the new anchor to predictor history; every cache step records the latest anchor as its reference, even when the reuse buffer contains a corrected value.

In the notation of Section~\ref{sec:online-correction}, this policy supplies the reference sequence $\br_t^{\mathrm{O}}=\boldsymbol{f}_t$.
Let $\boldsymbol{h}_{\tau-1}$ and $\boldsymbol{h}_{\tau-2}$ denote the two stored references preceding the current step.
At a first cache step, the reuse buffer still contains the newly refreshed full-compute anchor, so the three inputs to the second difference in Equation~(\ref{eq:acceleration}) are
\[
(\br_{\tau-2}^{\mathrm{O}},\br_{\tau-1}^{\mathrm{O}},\br_\tau^{\mathrm{O}})
=(\boldsymbol{h}_{\tau-2},\boldsymbol{h}_{\tau-1},\boldsymbol{f}_\tau).
\]
After computing the current correction, the state updates are
\begin{equation}
H_{\mathrm{pred}}\leftarrow\operatorname{append}
  (H_{\mathrm{pred}},\boldsymbol{f}_{\tau}),
\qquad
B_{\mathrm{reuse}}\leftarrow\widetilde{\br}_{\tau}.
\label{eq:isolated-state-update}
\end{equation}
Here $\operatorname{append}$ retains the two most recent reference entries.
Subsequent cache steps reuse the corrected residual until the next full evaluation, while predictor history continues to record the full anchor.
At these interior steps, the reference residuals repeat the same anchor and have zero second difference; no new correction is computed, and the existing corrected reuse value is retained.
The full anchors themselves evolve with the deployed rollout and are distinct from the offline full-compute trajectory.

In zero-based execution order, the shared schedules are:
\begin{itemize}[leftmargin=1.4em,itemsep=2pt,topsep=3pt]
\item $N=3$: full-compute steps $\{0,1,4,7,10,13,16,19\}$; correction steps $\{2,5,8,11,14,17\}$.
\item $N=4$: full-compute steps $\{0,1,2,6,10,14,18\}$; correction steps $\{3,7,11,15\}$, with step $19$ using direct reuse.
\item $N=5$: full-compute steps $\{0,1,2,7,12,17\}$; correction steps $\{3,8,13,18\}$.
\end{itemize}
Thus the three settings use eight, seven, and six full-compute steps, respectively, with all remaining steps using cached features.

\subsection{Calibration Budget and Parameter Footprint}
\label{app:calibration-cost}

Each deployment configuration uses one full-compute/naive-cache trajectory pair, followed by closed-form centered scalar fitting without iterative optimization.
The shared schedules activate six correction steps at $N=3$ and four at $N=4,5$.
FLUX.1-dev, Qwen-Image, and DiT-XL/2 use six, four, and two branch groups, respectively, with each scalar shared across layers and tokens.
Table~\ref{tab:calibration-footprint} counts the coefficients used at these correction steps and their FP32 payload, excluding serialization metadata and runtime feature buffers.

\begin{table}[t]
\centering
\caption{Active calibration coefficients per deployment configuration. Each entry gives the scalar count and its FP32 payload in bytes.}
\label{tab:calibration-footprint}
\small
\setlength{\tabcolsep}{7pt}
\begin{tabular}{l c c c}
\toprule
Model & $N=3$ & $N=4$ & $N=5$ \\
\midrule
FLUX.1-dev & 36 (144 B) & 24 (96 B) & 24 (96 B) \\
Qwen-Image & 24 (96 B) & 16 (64 B) & 16 (64 B) \\
DiT-XL/2 & 12 (48 B) & 8 (32 B) & 8 (32 B) \\
\bottomrule
\end{tabular}
\end{table}

For each deployment configuration, one trajectory pair supplies the sufficient statistics for closed-form scalar fitting, and the resulting maps remain fixed across evaluation samples.
The $8$--$36$ active scalars occupy $32$--$144$ bytes in FP32; this coefficient payload is distinct from runtime feature-history and reuse buffers.
After collection, each coefficient is obtained directly from accumulated inner products and squared norms, without iterative updates or a coefficient sweep.
These quantities characterize the calibration-data and parameter budgets.
Trajectory collection and fitting are one-time offline preparation, separate from both sampling warmup and benchmark execution warmup.
The reported denoising latency and FLOPs include the full deployment computation, including cache/history operations and additional network operations; offline preparation is excluded.

\subsection{Latency and FLOP Accounting}
\label{app:efficiency}

\paragraph{Denoising latency.}
All main-table measurements use a single NVIDIA H800 GPU and batch size one.
After ten execution warmup runs, we time 30 runs with CUDA events and report the mean per-image denoising latency.
Model weights and cache or calibration parameters are loaded onto the GPU before measurement.
The timed region includes the 20-step denoising loop, executed Transformer branches, cache/history operations, and any prediction or correction performed during denoising.
It excludes text encoding, VAE decoding, pack preparation, parameter loading, and disk I/O.
Speedup is the displayed full-compute baseline latency divided by the accelerated-method latency within the same model table.

\paragraph{FLOPs.}
FLOPs are measured with \texttt{torch.profiler} in a separate profiling run.
The reported main-table FLOPs for all three models, including the \method{} rows repeated in Table~\ref{tab:ablation-uncorrected-reuse}, use the same denoising-only boundary as latency: 20 sampling steps, excluding text encoding, VAE decoding, device-to-host transfer, and disk I/O.
We combine profiler-visible tensor operations with analytic counts for fused attention matrix multiplications, custom Triton correction kernels, and TaylorSeer/HiCache predictor arithmetic omitted by the profiler, avoiding double counting.
The parenthesized FLOPs reduction is the full-compute baseline total divided by the accelerated total.

\subsection{Correction-Target Study}
\label{app:mechanism-details}

Table~\ref{tab:ablation-correction-target} compares residual, velocity, and acceleration supervision while keeping the online update and deployment policy fixed.
The difference is how the correction coefficient is fitted offline.

\paragraph{Shared protocol.}
The study uses DiT-XL/2 on ImageNet-256.
All three variants use DDIM-20, $N=5$, warmup-3, CFG 1.5, one calibration trajectory, step-by-branch centered scalar fitting without a deployed intercept, and 50K generated images.
Their logical full-compute steps are $\{19,18,17,12,7,2\}$, and correction is applied only at logical steps $\{16,11,6,1\}$.
Every variant writes the corrected residual into the reuse buffer and retains full-anchor predictor-history isolation; subsequent cache steps reuse the stored corrected residual.
The acceleration row is the same deployed one-trajectory result as in Table~\ref{tab:dit}.

\paragraph{Shared online update.}
At a first cache step, let $\boldsymbol{f}$ be the latest full-compute residual and $\boldsymbol{p}$ the preceding anchor in the reference history of Appendix~\ref{app:dit-policy}.
Suppressing layer and branch indices, all three variants use
\begin{equation}
\widetilde{\br}_\tau
=\boldsymbol{f}+\alpha_g(\boldsymbol{f}-\boldsymbol{p}).
\label{eq:app-shared-predictor}
\end{equation}
The direction $\boldsymbol{f}-\boldsymbol{p}$ is the lagged difference between the two preceding reference entries.
The velocity variant scales this lagged difference. The current cached residual and the latest reference both equal $\boldsymbol{f}$, making the current difference zero.
For acceleration supervision, the three reference residuals are $(\boldsymbol{p},\boldsymbol{f},\boldsymbol{f})$, giving $\ba_\tau^{\mathrm{O}}=\boldsymbol{p}-\boldsymbol{f}$.
Substituting this into Equation~(\ref{eq:update}) gives
\begin{equation}
\begin{aligned}
\widetilde{\br}_\tau
&=\boldsymbol{f}+(s_g-1)(\boldsymbol{p}-\boldsymbol{f})\\
&=\boldsymbol{f}+(1-s_g)(\boldsymbol{f}-\boldsymbol{p}).
\end{aligned}
\label{eq:app-acceleration-coefficient}
\end{equation}
Thus $\alpha_g=1-s_g$ is the coefficient multiplying $\boldsymbol{f}-\boldsymbol{p}$ in the shared update; it is determined by the fitted $s_g$ and introduces no additional parameter.

\paragraph{Offline supervision.}
Table~\ref{tab:app-supervision-definitions} lists the fitting pairs and the resulting online coefficients.
Each row uses the centered objective and closed-form solution in Equations~(\ref{eq:ols}) and~(\ref{eq:closedform}), replacing the acceleration pair by $(\bx,\by)$.
The observations come from the independently evolving full-compute and naive-cache calibration trajectories.
We write $\boldsymbol{u}_\tau^{\cache}=\br_{\tau-1}^{\cache}-\br_{\tau-2}^{\cache}$ for the offline lagged difference; $\br_\tau^{\cache}$ is the residual reused at the current calibration step.
The fitted slopes are denoted by $\alpha_g^r$, $s_g^v$, and $s_g$ for residual, velocity, and acceleration supervision, respectively.

\begin{table}[t]
\centering
\caption{Supervision definitions for the common online update in Equation~(\ref{eq:app-shared-predictor}). The final column gives the coefficient multiplying $\boldsymbol{f}-\boldsymbol{p}$. Residual supervision targets the correction needed to match the full-trajectory residual. Velocity and acceleration supervision target the first and second residual differences along the full-compute trajectory.}
\label{tab:app-supervision-definitions}
\small
\setlength{\tabcolsep}{4pt}
\renewcommand{\arraystretch}{1.15}
\begin{tabular*}{\linewidth}{@{\extracolsep{\fill}}l l l c@{}}
\toprule
Supervision & Fit input $\bx$ & Fit target $\by$ & Online $\alpha_g$ \\
\midrule
Residual & $\boldsymbol{u}_\tau^{\cache}$ & $\br_\tau^{\full}-\br_\tau^{\cache}$ & $\alpha_g^r$ \\
Velocity & $\boldsymbol{u}_\tau^{\cache}$ & $\br_\tau^{\full}-\br_{\tau-1}^{\full}$ & $s_g^v$ \\
Acceleration & $\ba_\tau^{\cache}=-\boldsymbol{u}_\tau^{\cache}$ & $\ba_\tau^{\full}$ & $1-s_g$ \\
\bottomrule
\end{tabular*}
\end{table}

For acceleration supervision, $s_g=1$ gives direct reuse and $s_g=0$ gives extrapolation $2\boldsymbol{f}-\boldsymbol{p}$.
This two-anchor form is a linear predictor whose coefficient is estimated from paired cache-to-full acceleration observations; feature-reconstruction predictors such as L$^2$P~\citep{shen2026l2p} learn historical-feature weights from full-compute trajectories.
The results in Table~\ref{tab:ablation-correction-target} support acceleration supervision under this matched DiT protocol.

\paragraph{Separate residual-scaling control.}
Under the same calibration and deployment protocol, direct residual scaling fits $s_g^r$ offline with $\bx=\br_\tau^{\cache}$ and $\by=\br_\tau^{\full}$ and deploys $\widetilde{\br}_\tau=s_g^r\br_\tau^{\mathrm{O}}$ online.
It yields FID $38.17718$, sFID $30.14794$, IS $75.57263$, precision $0.44726$, and recall $0.45700$; its update form differs from the shared two-anchor predictor in Table~\ref{tab:ablation-correction-target}.

\paragraph{How acceleration discrepancies accumulate.}
An exact finite-difference identity relates acceleration discrepancies to residual error along the two offline calibration trajectories.
Define $\boldsymbol{e}_\tau=\br_\tau^{\cache}-\br_\tau^{\full}$, $\Delta\boldsymbol{e}_\tau=\boldsymbol{e}_\tau-\boldsymbol{e}_{\tau-1}$, and $\boldsymbol{\delta}_\tau=\ba_\tau^{\cache}-\ba_\tau^{\full}$.
Equation~(\ref{eq:acceleration}) then gives $\Delta^2\boldsymbol{e}_\tau=\boldsymbol{\delta}_\tau$.
Starting at a cache-interval boundary $\tau_0\geq1$ in execution order, summing twice yields, for $j\geq1$ and $\tau_0+j<T$,
\begin{equation}
\boldsymbol{e}_{\tau_0+j}
=\boldsymbol{e}_{\tau_0}+j\Delta\boldsymbol{e}_{\tau_0}
+\sum_{q=1}^{j}(j-q+1)\boldsymbol{\delta}_{\tau_0+q}.
\label{eq:app-accumulated-discrepancy}
\end{equation}
The first two terms carry the residual and first-difference errors at the boundary.
The sum includes every intervening step: earlier acceleration discrepancies receive larger weights because they contribute to more subsequent first differences.
If each discrepancy has norm at most $\epsilon$, the triangle inequality bounds the norm of this sum by $j(j+1)\epsilon/2$.
This is a worst-case bound on the accumulated contribution; vector terms may cancel.
A full evaluation on the cached rollout need not remove the boundary error, since its input can already differ from the full-compute trajectory.
The identity explains error accumulation. Table~\ref{tab:ablation-correction-target} supplies the empirical comparison.

\subsection{Writeback and Predictor-History Isolation}
\label{app:writeback-history}

The three policies in Table~\ref{tab:dit-writeback-history} differ in two states: the reuse buffer supplies features to the network, and predictor history supplies references for the next correction.
Each policy applies its correction at the eligible first cache step; full-compute steps refresh the reuse buffer and add a new full anchor to predictor history.

\paragraph{What writeback retains.}
Without writeback, the correction affects only the first cache step; later cache steps revert to the full-compute anchor.
Predictor history also records the full anchor at every cache step.
With writeback, later cache steps continue to use the corrected residual until the next full evaluation.
For either policy, let $\boldsymbol{f}$ denote its current full-compute anchor and $\boldsymbol{c}$ the correction computed at its first cache step.
Using an interval-local index with the full evaluation at $k=0$, both policies use $\boldsymbol{f}+\boldsymbol{c}$ at $k=1$.
At later cache steps before the next full evaluation at $k=N$,
\begin{equation}
\br_k^{\mathrm{used}}=
\begin{cases}
\boldsymbol{f},&\text{without writeback},\\
\boldsymbol{f}+\boldsymbol{c},&\text{with writeback},
\end{cases}
\qquad 2\leq k<N.
\label{eq:app-held-correction}
\end{equation}
Under isolated history, the reference entries repeat the same anchor after the first cache step, so their second difference is zero.
Writeback preserves the earlier correction even when no new correction is computed.
The terminal-step fallback in Appendix~\ref{app:deployment} still applies.

\paragraph{What history isolation changes.}
Under ordinary writeback, later cache steps use corrected residuals in both the reuse buffer and predictor history.
These corrected history entries can influence the correction in a subsequent interval.
Isolated history records the full anchor at every cache step, removing this direct feedback while the network continues to reuse corrected features.
To see the input difference at the next correction, let $\boldsymbol{p}$ be the preceding full anchor and $\boldsymbol{c}_{\mathrm{prev}}$ its retained correction, so the preceding interval ends with the reused residual $\boldsymbol{p}+\boldsymbol{c}_{\mathrm{prev}}$.
After a new full evaluation produces $\boldsymbol{f}$, the current residual at the first cache step is again $\boldsymbol{f}$.
Holding these anchors fixed to compare history handling, the acceleration inputs are
\begin{equation}
\begin{aligned}
\ba_\tau^{\mathrm{ord}}
&=\boldsymbol{f}-2\boldsymbol{f}+(\boldsymbol{p}+\boldsymbol{c}_{\mathrm{prev}})
=(\boldsymbol{p}-\boldsymbol{f})+\boldsymbol{c}_{\mathrm{prev}},\\
\ba_\tau^{\mathrm{iso}}
&=\boldsymbol{f}-2\boldsymbol{f}+\boldsymbol{p}
=\boldsymbol{p}-\boldsymbol{f}.
\end{aligned}
\label{eq:app-history-inputs}
\end{equation}
Here $\mathrm{ord}$ and $\mathrm{iso}$ denote ordinary writeback and isolated history.
Under Equation~(\ref{eq:update}), this additional history term contributes $(s_g-1)\boldsymbol{c}_{\mathrm{prev}}$ directly to the ordinary-writeback correction.
Isolation removes this term while retaining the corrected reuse buffer.
Consequently, the predictor's reference sequence and the residual sequence used by the network are distinct.
Though full anchors still evolve along the corrected rollout, the controlled comparison in Table~\ref{tab:dit-writeback-history} provides the empirical support for this policy.

\paragraph{Ablation implementation and protocol.}
The three runs share the same one-trajectory DiT scalar pack, full/cache/correction positions, two-rank sampling configuration, global seed zero, per-rank sampling batch size 128, and VAE decode batch size 8.
Each variant is evaluated on 50K generated images against the same ImageNet-256 reference.

\subsection{Fitting-Form Ablation}
\label{app:fitting-form-ablation}

We compare alternative scalar fitting forms for DiT-XL/2 on ImageNet-256 with DDIM-20, $N=5$, warmup-3, and 50K generated images. All variants use a single calibration trajectory, corrected-cache writeback, and full-anchor predictor-history isolation. The quantitative results are summarized in Table~\ref{tab:app-fitting-form-ablation}.
\begin{table*}[t]
\centering
\caption{Fitting-form ablation for DiT-XL/2 on ImageNet-256. Lower FID and sFID are better; higher IS, precision, and recall are better.}
\label{tab:app-fitting-form-ablation}
\small
\setlength{\tabcolsep}{5pt}
\begin{tabular}{l c c c c c}
\toprule
Fitting form & FID$\downarrow$ & sFID$\downarrow$ & IS$\uparrow$ & Precision$\uparrow$ & Recall$\uparrow$ \\
\midrule
Centered $s$-only ($b=0$) & 10.36915 & 6.40612 & 151.75644 & 0.66828 & 0.57380 \\
Through-origin slope-only & 14.18379 & 8.65522 & 136.35312 & 0.62440 & 0.58140 \\
$s{+}b$ & 10.35668 & 6.33262 & 151.42812 & 0.66792 & 0.56760 \\
\bottomrule
\end{tabular}
\end{table*}

In this DiT setting, centered $s$-only fitting improves FID, sFID, IS, and precision over through-origin fitting. Compared with $s{+}b$, the scalar-only deployed map achieves similar FID and sFID and higher IS, precision, and recall.

\Needspace{34\baselineskip}
\subsection{Coefficient Choice and Sharing}
\label{app:coefficient-choice}

We compare coefficient settings on FLUX.1-dev, DrawBench-200, with 20 sampling steps, $N=5$, warmup-3, and $1024^2$ resolution.
In the two-anchor form of Appendix~\ref{app:mechanism-details}, fixing $s=0$ gives extrapolation $\widetilde{\br}_\tau=2\boldsymbol{f}-\boldsymbol{p}$, whereas $s=1$ reduces algebraically to direct reuse.
The matched-schedule Uncorrected reuse comparison is reported in Table~\ref{tab:ablation-uncorrected-reuse}.
Table~\ref{tab:coefficient-choice} compares fixed extrapolation, a shared fitted scalar $s=0.26131$, and the step-by-branch coefficients used by \method{}.
The shared and grouped fits use the same calibration trajectory pair (COCO prompt-file index 50000, noise seed 364159).
We constrain the 24 active group slopes to one value using their stored sufficient statistics, retaining channelwise centering within each original group:
\[
s_{\mathrm{shared}}=\frac{\sum_{g=1}^{24}C_g}{\sum_{g=1}^{24}D_g}\approx0.26131,
\]
where $C_g$ and $D_g$ are the centered cross-product and input-energy sums, respectively, with the original per-channel denominator floor of $10^{-12}$.
The summed objective uses no additional group weighting or normalization, and the coefficient is estimated independently of the evaluation results.
All three settings use the same correction positions, corrected-cache writeback, isolated full-anchor history, branch projections, inference implementation, and evaluation prompts and seeds; only the active coefficient values change.
The grouped row reuses the main FLUX $N=5$ result.

\begin{table}[!htbp]
\centering
\caption{Coefficient-choice ablation on FLUX.1-dev, DrawBench-200, with 20 steps, $N=5$, and warmup-3. The scalar count denotes independently fitted parameters; bold marks the best reported mean.}
\label{tab:coefficient-choice}
\small
\setlength{\tabcolsep}{3pt}
\renewcommand{\arraystretch}{1.08}
\begin{tabular*}{\linewidth}{@{\extracolsep{\fill}}l r r r r r r@{}}
\toprule
Coefficient setting & \shortstack{Fitted\\scalars} & PSNR$\uparrow$ & SSIM$\uparrow$ & LPIPS$\downarrow$ & CLIP$\uparrow$ & IR$\uparrow$ \\
\midrule
Fixed $s=0$ & 0 & 17.3736 & 0.59297 & 0.57690 & 31.8232 & 0.79039 \\
Shared $s=0.26131$ & 1 & 18.4725 & 0.67029 & 0.43009 & 31.8310 & 0.87135 \\
Grouped $s_g$ (\method{}) & 24 & \best{19.1014} & \best{0.67348} & \best{0.42925} & \best{31.9461} & \best{0.89891} \\
\bottomrule
\end{tabular*}
\end{table}

Both fitted settings outperform fixed extrapolation in the reported aggregates. Grouping further improves PSNR by $0.6289$ dB over the shared fit, while the shared scalar yields similar SSIM and LPIPS.

\subsection{Additional Calibration Ablations}
\label{app:additional-calibration-ablations}

This subsection examines sensitivity to calibration-set size and the identity of a single calibration prompt. The matched-schedule correction comparison is reported in Table~\ref{tab:ablation-uncorrected-reuse} in the main text.

\paragraph{Number of calibration prompts.}
We evaluate calibration budgets of 1, 4, 16, and 64 nested COCO prompts on FLUX.1-dev with 20 steps, $N=5$, and warmup-3, using the same DrawBench-200 evaluation prompts and seeds.
Each prompt supplies one paired full-compute/naive-cache trajectory; all packs use centered $s$-only fitting, step-by-branch sharing, corrected-cache writeback, and isolated full-anchor predictor history.
The one-prompt setting uses the current main-table pack. Table~\ref{tab:ablation-calibration-count} reports the completed four-setting study.
Increasing the number of calibration trajectories does not yield a clear quality benefit.

\begin{table}[t]
\centering
\caption{Calibration-set-size sensitivity for FLUX.1-dev with 20 steps, $N=5$, warmup-3, and step-by-branch sharing on DrawBench-200. Each calibration prompt supplies one trajectory pair.}
\label{tab:ablation-calibration-count}
\small
\setlength{\tabcolsep}{5pt}
\begin{tabular}{r r r r r r}
\toprule
Calib. prompts & PSNR$\uparrow$ & SSIM$\uparrow$ & LPIPS$\downarrow$ & CLIP$\uparrow$ & IR$\uparrow$ \\
\midrule
1  & 19.1014 & 0.67348 & 0.42925 & 31.9461 & 0.89891 \\
4  & 19.2467 & 0.68151 & 0.41684 & 31.8786 & 0.90214 \\
16 & 19.1494 & 0.67215 & 0.43108 & 31.8759 & 0.88410 \\
64 & 19.1270 & 0.66995 & 0.43479 & 31.8941 & 0.88120 \\
\bottomrule
\end{tabular}
\end{table}

\paragraph{Random single-prompt calibration.}
To test sensitivity to prompt identity, we independently fit ten step-by-branch scalar packs by centered least squares, each on one randomly selected COCO prompt disjoint from DrawBench-200. The selection seed is fixed in advance, and all ten fits use the same calibration noise seed and unified deployment policy. Every pack is evaluated on the same 200 DrawBench prompts and evaluation seeds. Table~\ref{tab:ablation-random-one} reports the mean, sample standard deviation, and range of the resulting DrawBench-200 aggregate for each metric across the ten packs. The small dispersion across packs indicates limited sensitivity to the identity of the single calibration prompt in this setting.

\begin{table}[t]
\centering
\caption{Random single-prompt calibration on FLUX.1-dev, DrawBench-200, $N=5$, warmup-3. Each of ten independently fitted step-by-branch packs uses the same calibration noise seed; evaluation prompts and seeds are shared. Mean, sample standard deviation, and range are across the ten pack-level aggregates.}
\label{tab:ablation-random-one}
\small
\setlength{\tabcolsep}{6pt}
\renewcommand{\arraystretch}{1.10}
\begin{tabular}{@{}l r r@{}}
\toprule
Metric & Mean $\pm$ std & Range \\
\midrule
CLIP$\uparrow$ & $31.9113 \pm 0.0255$ & $[31.8755,\ 31.9534]$ \\
IR$\uparrow$ & $0.89312 \pm 0.01033$ & $[0.87758,\ 0.90675]$ \\
PSNR$\uparrow$ & $19.1151 \pm 0.0848$ & $[18.9346,\ 19.2567]$ \\
SSIM$\uparrow$ & $0.67070 \pm 0.00673$ & $[0.65619,\ 0.68096]$ \\
LPIPS$\downarrow$ & $0.43368 \pm 0.00956$ & $[0.41778,\ 0.45368]$ \\
\bottomrule
\end{tabular}
\end{table}

\subsection{Additional Comparison with L\texorpdfstring{$^2$}{2}P}
\label{app:l2p-comparison}

We additionally compare \method{} with L$^2$P~\citep{shen2026l2p}, a data-driven linear predictor that learns timestep-specific combinations of historical features from full-compute trajectories.
This comparison is placed in the appendix because the publicly released L$^2$P implementation does not support experiments on all three models evaluated in the main text, so we developed our own implementation.
However, we find that our adaptation of L$^2$P performs very poorly on DiT.

\paragraph{Adaptation and evaluation protocol.}
The released implementation does not provide DiT support, so the results below therefore use our adaptation of its output-level predictor to DiT-XL/2.
We retain its causal lower-triangular weight matrix, identity initialization, masked mean-squared-error objective, and Adam optimization with learning rate $0.01$ for $201$ updates.
The predictor is fitted on $64$ full-compute trajectories, with $16$ additional trajectories used only for validation monitoring.
At each cache step, it predicts the complete sampler-facing model output from preceding full or predicted outputs and appends the prediction to its history.
This preserves L$^2$P's output-level design: one prediction per cache step avoids maintaining and predicting attention/MLP features at every Transformer layer, keeping prediction and history-storage overhead low.
We report the best-performing completed adaptation.

Both methods use ImageNet-256, DDIM-20, CFG $1.5$, and the same $N=5$/warmup-3 schedule, with logical full-compute steps $\{19,18,17,12,7,2\}$.
Quality is evaluated on $50$K generated images.
Latency and FLOPs follow the denoising-only accounting in Appendix~\ref{app:efficiency}; latency uses one H800, batch size one, and $30$ CUDA-event measurements after $10$ execution warmup runs.

\begin{table*}[t]
\centering
\caption{Comparison with our L$^2$P adaptation on DiT-XL/2, ImageNet-256, using DDIM-20, CFG $1.5$, $N=5$, warmup-3, and $50$K generated images. Traj. counts calibration conditioning/noise realizations: L$^2$P uses $64$ full-compute trajectories, whereas \method{} uses one full-compute/naive-cache trajectory pair. Coeff. counts active deployed coefficients.}
\label{tab:l2p-dit}
\scriptsize
\setlength{\tabcolsep}{3pt}
\begin{tabular*}{\textwidth}{@{\extracolsep{\fill}}l r r r r r r r r r@{}}
\toprule
Method & Traj. & Coeff. & Latency (s)$\downarrow$ & FLOPs (T)$\downarrow$ & FID$\downarrow$ & sFID$\downarrow$ & IS$\uparrow$ & Precision$\uparrow$ & Recall$\uparrow$\\
\midrule
L$^2$P (adapted) & 64 & 151 & \textbf{0.0829} & \textbf{2.7461} & 39.9860 & 28.7022 & 79.2690 & 0.47538 & 0.54490\\
\method{} & \textbf{1} & \textbf{8} & 0.1354 & 2.7622 & \textbf{10.3691} & \textbf{6.4061} & \textbf{151.7564} & \textbf{0.66828} & \textbf{0.57380}\\
\bottomrule
\end{tabular*}
\end{table*}

Table~\ref{tab:l2p-dit} confirms the latency advantage of output-level prediction, but also shows a large quality gap: L$^2$P obtains FID $39.9860$ and sFID $28.7022$, compared with $10.3691$ and $6.4061$ for \method{}.
\method{} improves all five quality metrics while using one calibration trajectory pair and $8$ deployed coefficients, compared with $64$ trajectories and $151$ coefficients for L$^2$P.

\paragraph{Mismatch between fitting and deployment histories.}
L$^2$P is fitted using true historical outputs from full-compute trajectories, whereas its deployment history also contains previously predicted outputs.
Let $f_k$ denote the full-compute reference output at execution index $k$, let $\widetilde f_k$ denote the deployed output, and define $e_k=\widetilde f_k-f_k$.
At a cache step, the prediction error satisfies
\begin{equation}
e_k
=\underbrace{\left(\sum_{j<k}w_{k,j}f_j-f_k\right)}_{\text{true-history prediction residual}}
+\underbrace{\sum_{j<k}w_{k,j}e_j}_{\text{propagated history errors}}.
\label{eq:l2p-history-mismatch}
\end{equation}
Thus, a small prediction residual under true history does not guarantee a small deployed error.
Prediction errors also alter the latent trajectory, so later full-compute outputs can differ from the reference even when those steps are evaluated exactly.

To test the effect of history mismatch, we keep the fitted weights fixed and continue to use predicted outputs for sampling at cache steps, but populate the predictor history with true model outputs evaluated on the current deployed trajectory.
Across $32$ paired samples from two random seeds, this diagnostic reduces final latent MSE relative to full computation from $0.08146$ to $0.02995$, a $63.2\%$ reduction.

Our reproduction of L$^2$P performs poorly on DiT, a discrete diffusion model, for which no official implementation is publicly available. Moreover, L$^2$P primarily reports acceleration results under 50-step sampling, whereas our study focuses on shorter sampling schedules, particularly 20 steps. Given these limitations, we do not consider our reproduction sufficiently validated to serve as a reliable baseline in the main text. 

\subsection{Additional Comparison with TC-Pad\'e}
\label{app:tcpade-comparison}

We additionally reproduce TC-Pad\'e~\citep{cui2026tcpade} from the authors' public repository (commit \texttt{e2d6302}) on the same FLUX.1-dev DrawBench-200 evaluation used in the main text.
The released repository snapshot contains executable inference code for FLUX.1-dev, but not corresponding implementations for Qwen-Image or DiT-XL/2.
TC-Pad\'e therefore cannot be evaluated under the main text's cross-model protocol, so we report the available FLUX-specific comparison in the appendix rather than adding it to the main benchmark tables.
Our evaluation wrapper retains the released Transformer forward and residual predictor without modifying the repository source, while providing the current Diffusers interface, indexed prompt seeds, and per-prompt history reset.
All rows in Table~\ref{tab:tcpade-comparison} use 20 denoising steps, $1024^2$ resolution, guidance 3.5, and the same 200 prompts and seeds.
For TC-Pad\'e, $N_{\mathrm{th}}\in\{0.7,1.0,1.3\}$ denotes the repository's curvature threshold, where a larger value permits more aggressive prediction; it is not the fixed cache interval $N$ used by \method{}.
The reproduced configurations use predictor order 3, history size 6, and interval 8; prediction starts at step 4 for $N_{\mathrm{th}}=0.7,1.0$ and step 3 for $N_{\mathrm{th}}=1.3$.
Latency uses one H800 at batch size one, with 30 CUDA-event measurements after 10 warmup runs; FLOPs combine profiler-visible operations with analytic fused-attention matrix multiplications.
Both measurements cover the 20-step denoising process only and exclude text encoding, VAE decoding, device-to-host transfer, and I/O, matching the main-table accounting boundary.
Because TC-Pad\'e makes prompt-dependent prediction decisions, its FLOPs are measured on one warmed trajectory using the first DrawBench prompt and its fixed evaluation seed.

\begin{table}[t]
\centering
\caption{Efficiency and quality comparison with our reproduction of the public TC-Pad\'e implementation on FLUX.1-dev, DrawBench-200, with 20 denoising steps at $1024^2$. TC-Pad\'e's $N_{\mathrm{th}}$ is a curvature threshold, whereas \method{}'s $N$ is a fixed cache interval.}
\label{tab:tcpade-comparison}
\scriptsize
\setlength{\tabcolsep}{2.5pt}
\begin{tabular*}{\linewidth}{@{\extracolsep{\fill}}l c c c c c c c@{}}
\toprule
Method & Latency (s)$\downarrow$ & FLOPs (T)$\downarrow$ & CLIP$\uparrow$ & IR$\uparrow$ & PSNR$\uparrow$ & SSIM$\uparrow$ & LPIPS$\downarrow$ \\
\midrule
TC-Pad\'e ($N_{\mathrm{th}}=0.7$) & 5.8379 & 1116.023 & 31.6134 & 0.89883 & 11.7447 & 0.52002 & 0.62702 \\
TC-Pad\'e ($N_{\mathrm{th}}=1.0$) & 3.4730 & 669.747 & 31.5535 & 0.82747 & 11.8787 & 0.53759 & 0.65077 \\
TC-Pad\'e ($N_{\mathrm{th}}=1.3$) & 3.0231 & 520.987 & 31.4045 & 0.69882 & 11.9939 & 0.54075 & 0.66214 \\
\midrule
\method{} ($N=3$) & 3.2094 & 621.914 & 31.5376 & 0.95549 & 18.6202 & 0.72185 & 0.34093 \\
\method{} ($N=4$) & 2.8173 & 544.223 & 31.6116 & \best{0.96672} & \best{19.8328} & \best{0.74463} & \best{0.28787} \\
\method{} ($N=5$) & \best{2.4644} & \best{466.550} & \best{31.9461} & 0.89891 & 19.1014 & 0.67348 & 0.42925 \\
\bottomrule
\end{tabular*}
\end{table}

Across the displayed configurations, \method{} gives substantially higher full-compute fidelity than the reproduced TC-Pad\'e configurations in PSNR, SSIM, and LPIPS, and also improves ImageReward in the matched DrawBench evaluation.
CLIP is closer and configuration-dependent: TC-Pad\'e at $N_{\mathrm{th}}=0.7$ is slightly higher than \method{} at $N=3,4$, while \method{} at $N=5$ obtains the highest CLIP score in this comparison.
More aggressive TC-Pad\'e thresholds reduce latency and FLOPs, while \method{} at $N=5$ remains faster and uses fewer FLOPs than all three reproduced TC-Pad\'e configurations.

\clearpage
\section{Additional Qualitative Comparisons}
\label{app:qualitative-comparisons}

\begin{figure*}[!h]
  \centering
  \includegraphics[width=\textwidth,height=0.74\textheight,keepaspectratio]{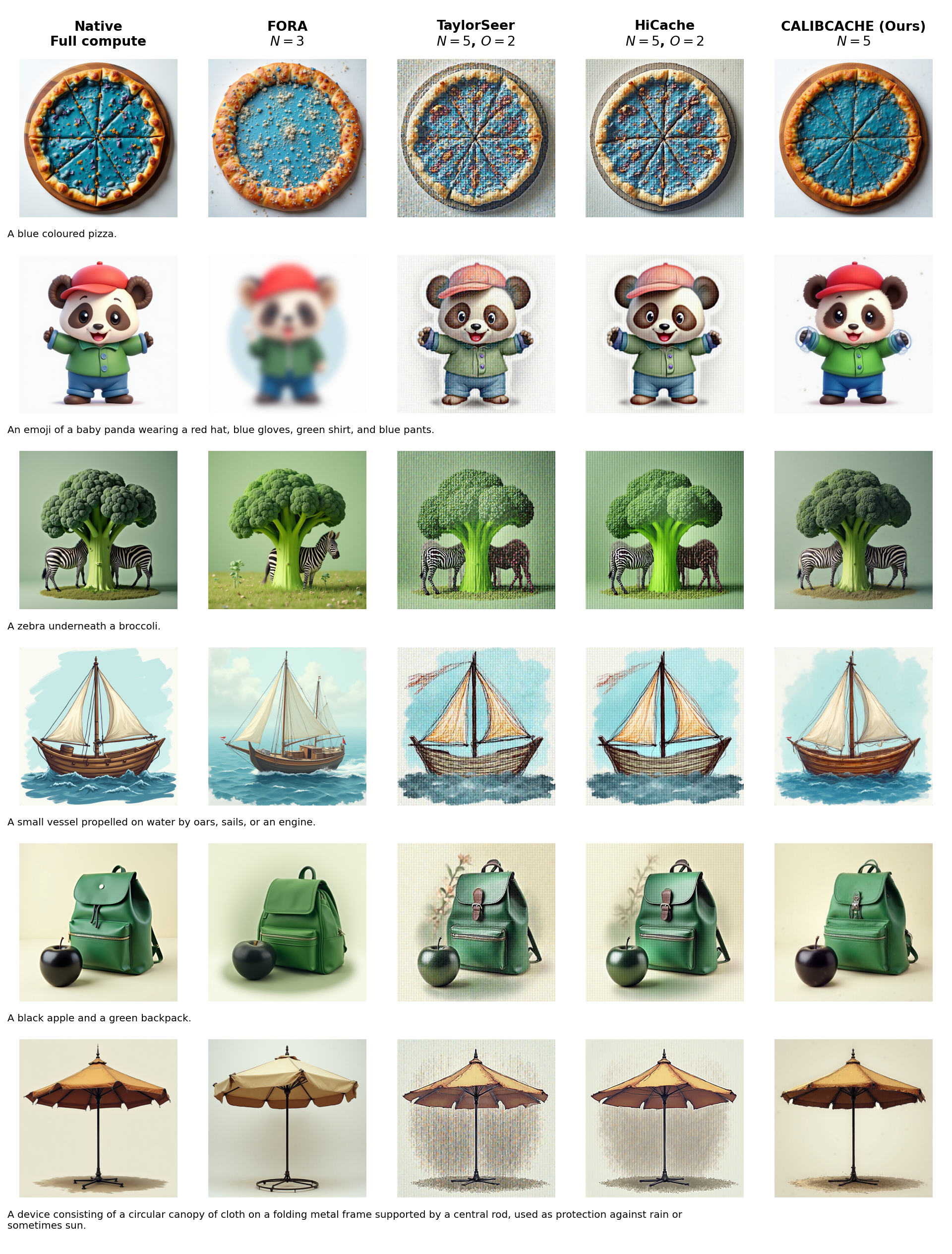}
  \caption{FLUX.1-dev comparisons on DrawBench at $1024^2$ with 20 denoising steps. Six prompts are shown in rows, with the exact prompt below each row. As in the introductory figure, FORA uses $N=3$ without extra warmup; TaylorSeer, HiCache, and \method{} use $N=5$ and warmup-3. TaylorSeer and HiCache use $O=2$, and HiCache uses scale $0.6$. These examples highlight reduced texture artifacts and closer preservation of the full-compute appearance and layout.}
  \label{fig:app-qualitative-flux}
\end{figure*}

\begin{figure*}[p]
  \centering
  \includegraphics[width=\textwidth,height=0.85\textheight,keepaspectratio]{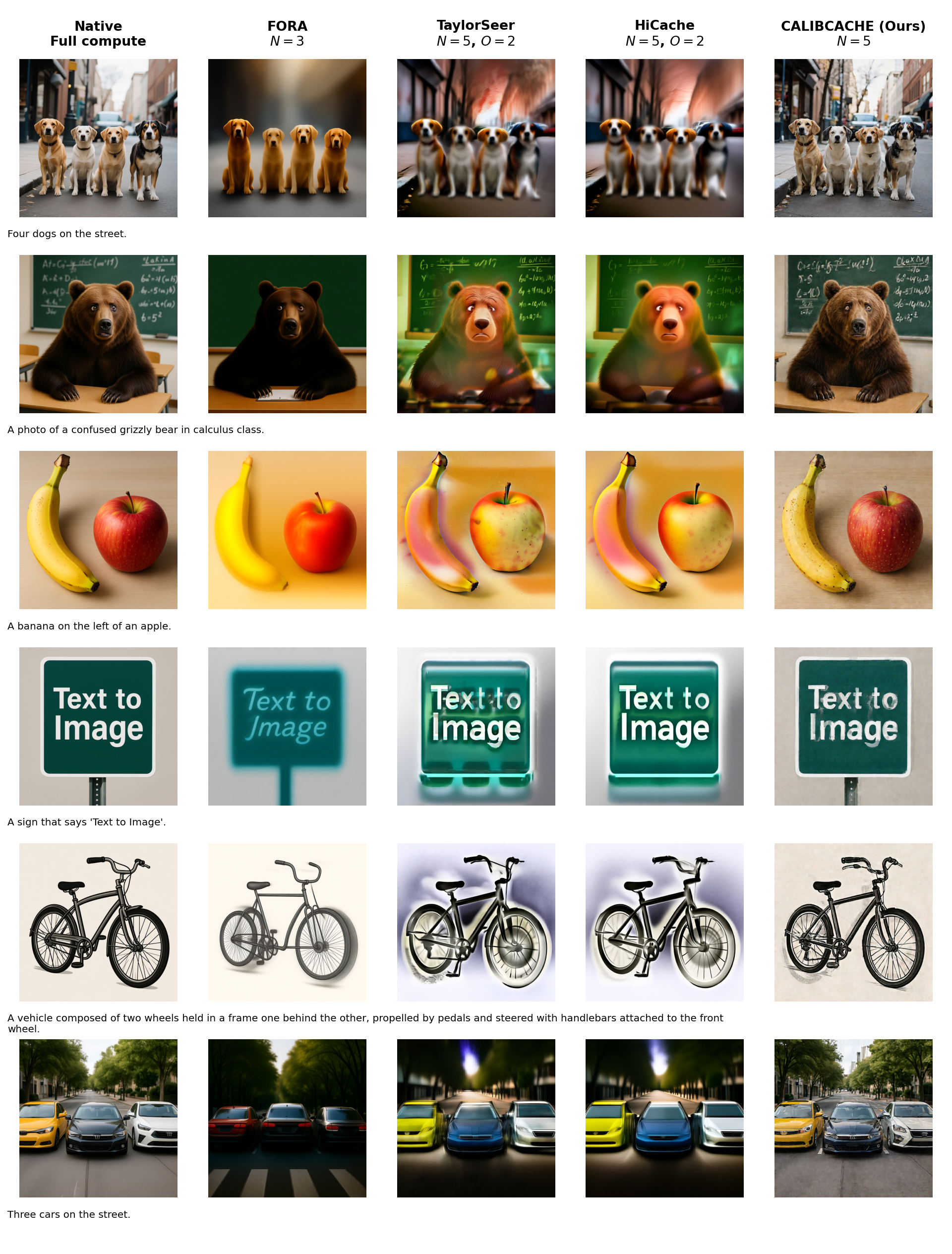}
  \caption{Qwen-Image comparisons on DrawBench at $1024^2$ with 20 denoising steps. FORA uses $N=3$ without extra warmup; TaylorSeer, HiCache, and \method{} use $N=5$ and warmup-3. TaylorSeer and HiCache use $O=2$, and HiCache uses scale $0.6$. These examples illustrate clearer subjects and backgrounds and closer preservation of the full-compute scene and material appearance.}
  \label{fig:app-qualitative-qwen}
\end{figure*}

\begin{figure*}[p]
  \centering
  \includegraphics[width=\textwidth,height=0.82\textheight,keepaspectratio]{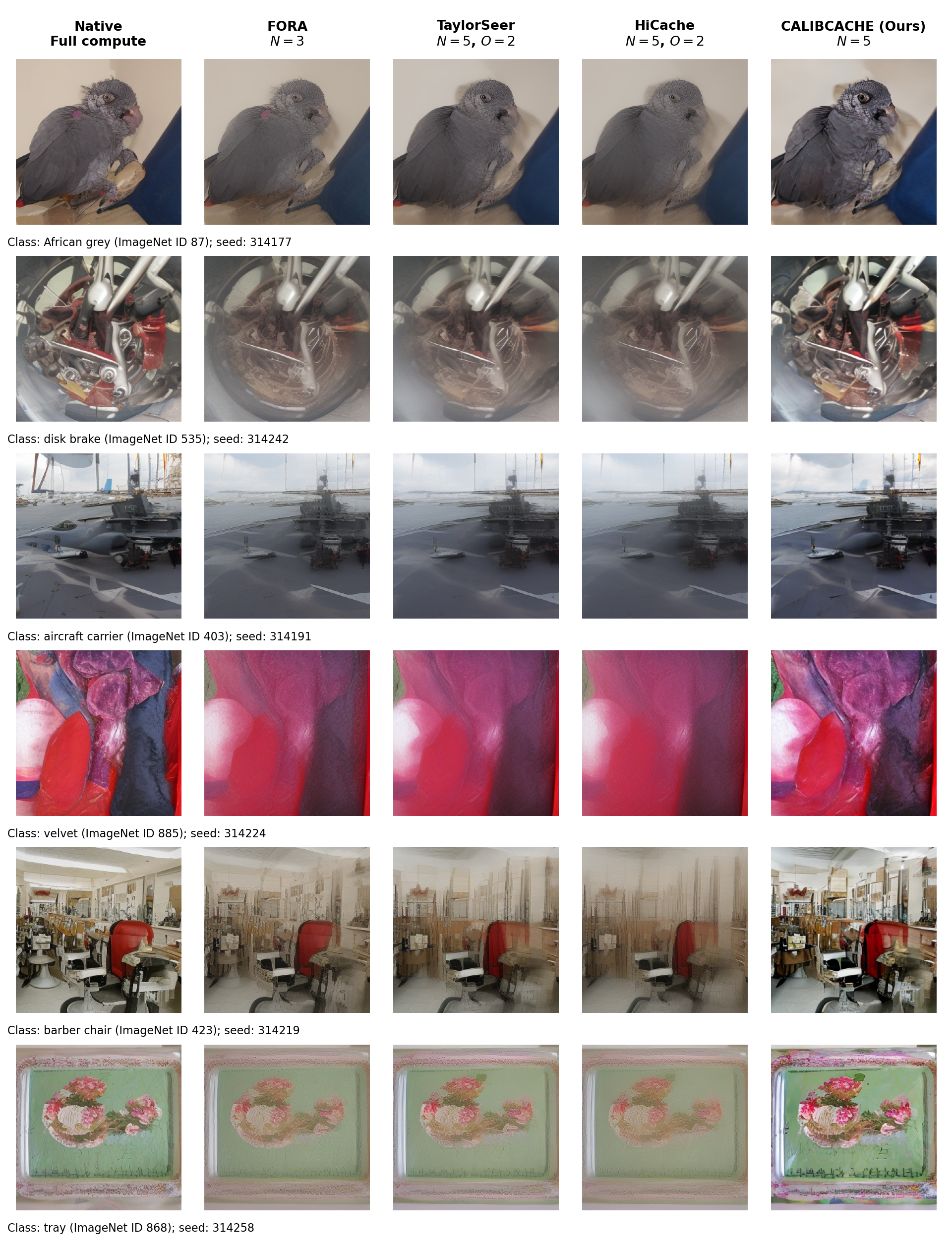}
  \caption{DiT-XL/2 comparisons on ImageNet-256 with DDIM-20 and CFG 1.5. Six examples are selected from 100 paired class--noise inputs, with the class and noise seed below each row. All methods start from the identical saved latent and class label for each example. FORA uses $N=3$ without extra warmup; TaylorSeer, HiCache, and \method{} use $N=5$ and warmup-3. TaylorSeer and HiCache use $O=2$, and HiCache uses scale $0.6$. These examples illustrate differences in texture and detail; aggregate quality is reported in Table~\ref{tab:dit}.}
  \label{fig:app-qualitative-dit}
\end{figure*}

Figures~\ref{fig:app-qualitative-flux}, \ref{fig:app-qualitative-qwen}, and~\ref{fig:app-qualitative-dit} show more appearance and detail comparisons using matched conditioning and initial noise; aggregate quality is reported in the quantitative tables.
All \method{} examples use one-trajectory centered scalar fitting, corrected-cache writeback, and isolated full-anchor history, without image retouching or method-specific enhancement.

\clearpage

\clearpage

\clearpage

\section{Evaluation at Longer Sampling Schedules}
\label{app:longer-schedules}
\textbf{Evaluation protocol.}
We extend the text-to-image evaluation to 30, 40, and 50 sampling steps on
FLUX.1-dev and Qwen-Image. Each configuration uses the same 200 DrawBench
prompts at $1024\times1024$, batch size one, and paired evaluation seeds
$314159+i$ for prompt index $i\in\{0,\ldots,199\}$.
Guidance is 3.5 for FLUX and true CFG 4.0 for Qwen-Image.
Each model and sampling grid uses one full-compute/naive-cache calibration
trajectory pair from a calibration prompt disjoint from the evaluation set; the frozen coefficients are shared across the
evaluation set. Fidelity metrics use the full-compute output at the same grid,
prompt, and seed as the reference.

\textbf{Cache schedules.}
All caching methods use $N=5$. TaylorSeer, HiCache, and CalibCache use three
initial full-compute steps and then refresh every five steps, giving 8, 10,
and 12 full-compute steps at 30, 40, and 50 steps, respectively. FORA uses its
standard schedule without extra warmup, giving 6, 8, and 10 full-compute steps.
TaylorSeer and HiCache use order two; HiCache uses scale 0.6.
CalibCache retains corrected-cache writeback and isolated full-anchor history
and skips the last correction at zero-based indices 28, 38, and 48,
respectively. These are sampling-step counts; Qwen evaluates both CFG branches.

\textbf{FLUX.1-dev results.}
Tables 16--18 report all metrics for the three sampling grids.
Across Tables 16--21, bold values denote the best result among caching methods; full-compute reference rows are excluded from ranking.
\par
\begin{table}[!ht]
\centering
\caption{FLUX.1-dev with 30 sampling steps on DrawBench-200. IR denotes ImageReward; latency is denoising seconds per image.}
\label{tab:longgrid-flux-30}
\vspace{2pt}\centering\fontsize{9}{10.8}\selectfont
\setlength{\tabcolsep}{2.0pt}
\begin{tabular*}{\linewidth}{@{\extracolsep{\fill}}lrrrrrrr@{}}
\toprule
Method & Latency$\downarrow$ & Speedup$\uparrow$ & PSNR$\uparrow$ & SSIM$\uparrow$ & LPIPS$\downarrow$ & CLIP$\uparrow$ & IR$\uparrow$ \\
\midrule
FLUX.1-dev (30 steps) & 10.225 & 1.00$\times$ & $\infty$ & 1.0000 & 0.0000 & 31.236 & 0.9937 \\
FORA & \textbf{2.347} & \textbf{4.36$\times$} & 14.42 & 0.5801 & 0.5577 & \textbf{31.433} & 0.6679 \\
TaylorSeer & 3.530 & 2.90$\times$ & 17.07 & 0.6864 & 0.3637 & 31.300 & 0.9526 \\
HiCache & 3.435 & 2.98$\times$ & 17.50 & 0.7056 & 0.3365 & 31.314 & 0.9495 \\
\textsc{CalibCache} (Ours) & 3.154 & 3.24$\times$ & \textbf{18.16} & \textbf{0.7060} & \textbf{0.3192} & 31.379 & \textbf{0.9689} \\
\bottomrule
\end{tabular*}
\end{table}
\par
\begin{table}[!ht]
\centering
\caption{FLUX.1-dev with 40 sampling steps on DrawBench-200. IR denotes ImageReward; latency is denoising seconds per image.}
\label{tab:longgrid-flux-40}
\vspace{2pt}\centering\fontsize{9}{10.8}\selectfont
\setlength{\tabcolsep}{2.0pt}
\begin{tabular*}{\linewidth}{@{\extracolsep{\fill}}lrrrrrrr@{}}
\toprule
Method & Latency$\downarrow$ & Speedup$\uparrow$ & PSNR$\uparrow$ & SSIM$\uparrow$ & LPIPS$\downarrow$ & CLIP$\uparrow$ & IR$\uparrow$ \\
\midrule
FLUX.1-dev (40 steps) & 13.803 & 1.00$\times$ & $\infty$ & 1.0000 & 0.0000 & 31.263 & 0.9844 \\
FORA & \textbf{3.162} & \textbf{4.36$\times$} & 14.24 & 0.5993 & 0.5179 & \textbf{31.562} & 0.8264 \\
TaylorSeer & 4.564 & 3.02$\times$ & 16.81 & 0.6935 & 0.3607 & 31.370 & 0.9896 \\
HiCache & 4.446 & 3.10$\times$ & 17.15 & 0.7040 & 0.3371 & 31.361 & 0.9913 \\
\textsc{CalibCache} (Ours) & 4.029 & 3.43$\times$ & \textbf{17.78} & \textbf{0.7068} & \textbf{0.3264} & 31.511 & \textbf{1.0034} \\
\bottomrule
\end{tabular*}
\end{table}
\par
\begin{table}[!ht]
\centering
\caption{FLUX.1-dev with 50 sampling steps on DrawBench-200. IR denotes ImageReward; latency is denoising seconds per image.}
\label{tab:longgrid-flux-50}
\vspace{2pt}\centering\fontsize{9}{10.8}\selectfont
\setlength{\tabcolsep}{2.0pt}
\begin{tabular*}{\linewidth}{@{\extracolsep{\fill}}lrrrrrrr@{}}
\toprule
Method & Latency$\downarrow$ & Speedup$\uparrow$ & PSNR$\uparrow$ & SSIM$\uparrow$ & LPIPS$\downarrow$ & CLIP$\uparrow$ & IR$\uparrow$ \\
\midrule
FLUX.1-dev (50 steps) & 16.913 & 1.00$\times$ & $\infty$ & 1.0000 & 0.0000 & 31.215 & 0.9799 \\
FORA & \textbf{3.886} & \textbf{4.35$\times$} & 14.10 & 0.6048 & 0.4993 & 31.311 & 0.9097 \\
TaylorSeer & 5.490 & 3.08$\times$ & 16.52 & 0.6908 & 0.3620 & 31.324 & 0.9902 \\
HiCache & 5.350 & 3.16$\times$ & 16.86 & 0.7046 & 0.3392 & 31.277 & 1.0003 \\
\textsc{CalibCache} (Ours) & 4.781 & 3.54$\times$ & \textbf{17.46} & \textbf{0.7047} & \textbf{0.3334} & \textbf{31.535} & \textbf{1.0029} \\
\bottomrule
\end{tabular*}
\end{table}
\clearpage

\textbf{Qwen-Image results.}
Tables 19--21 report the corresponding evaluation on Qwen-Image.
\par
\begin{table}[!ht]
\centering
\caption{Qwen-Image with 30 sampling steps on DrawBench-200. IR denotes ImageReward; latency is denoising seconds per image.}
\label{tab:longgrid-qwen-30}
\vspace{2pt}\centering\fontsize{9}{10.8}\selectfont
\setlength{\tabcolsep}{2.0pt}
\begin{tabular*}{\linewidth}{@{\extracolsep{\fill}}lrrrrrrr@{}}
\toprule
Method & Latency$\downarrow$ & Speedup$\uparrow$ & PSNR$\uparrow$ & SSIM$\uparrow$ & LPIPS$\downarrow$ & CLIP$\uparrow$ & IR$\uparrow$ \\
\midrule
Qwen-Image (30 steps) & 19.382 & 1.00$\times$ & $\infty$ & 1.0000 & 0.0000 & 33.001 & 1.2189 \\
FORA & \textbf{6.960} & \textbf{2.78$\times$} & 12.57 & 0.4858 & 0.5882 & 30.953 & 0.1195 \\
TaylorSeer & 9.196 & 2.11$\times$ & 13.58 & 0.5817 & 0.4942 & 31.353 & 0.8149 \\
HiCache & 10.274 & 1.89$\times$ & 13.64 & 0.5779 & 0.4734 & 31.648 & 0.8857 \\
\textsc{CalibCache} (Ours) & 8.837 & 2.19$\times$ & \textbf{18.21} & \textbf{0.7122} & \textbf{0.3128} & \textbf{32.558} & \textbf{1.1231} \\
\bottomrule
\end{tabular*}
\end{table}
\par
\begin{table}[!ht]
\centering
\caption{Qwen-Image with 40 sampling steps on DrawBench-200. IR denotes ImageReward; latency is denoising seconds per image.}
\label{tab:longgrid-qwen-40}
\vspace{2pt}\centering\fontsize{9}{10.8}\selectfont
\setlength{\tabcolsep}{2.0pt}
\begin{tabular*}{\linewidth}{@{\extracolsep{\fill}}lrrrrrrr@{}}
\toprule
Method & Latency$\downarrow$ & Speedup$\uparrow$ & PSNR$\uparrow$ & SSIM$\uparrow$ & LPIPS$\downarrow$ & CLIP$\uparrow$ & IR$\uparrow$ \\
\midrule
Qwen-Image (40 steps) & 25.596 & 1.00$\times$ & $\infty$ & 1.0000 & 0.0000 & 32.970 & 1.2359 \\
FORA & \textbf{9.333} & \textbf{2.74$\times$} & 12.51 & 0.4906 & 0.5624 & 31.584 & 0.4950 \\
TaylorSeer & 11.951 & 2.14$\times$ & 14.48 & 0.6356 & 0.4155 & 32.355 & 1.0860 \\
HiCache & 13.438 & 1.90$\times$ & 14.53 & 0.6283 & 0.3977 & 32.407 & 1.0993 \\
\textsc{CalibCache} (Ours) & 11.556 & 2.21$\times$ & \textbf{18.32} & \textbf{0.7370} & \textbf{0.2595} & \textbf{32.982} & \textbf{1.2094} \\
\bottomrule
\end{tabular*}
\end{table}
\par
\begin{table}[!ht]
\centering
\caption{Qwen-Image with 50 sampling steps on DrawBench-200. IR denotes ImageReward; latency is denoising seconds per image.}
\label{tab:longgrid-qwen-50}
\vspace{2pt}\centering\fontsize{9}{10.8}\selectfont
\setlength{\tabcolsep}{2.0pt}
\begin{tabular*}{\linewidth}{@{\extracolsep{\fill}}lrrrrrrr@{}}
\toprule
Method & Latency$\downarrow$ & Speedup$\uparrow$ & PSNR$\uparrow$ & SSIM$\uparrow$ & LPIPS$\downarrow$ & CLIP$\uparrow$ & IR$\uparrow$ \\
\midrule
Qwen-Image (50 steps) & 31.723 & 1.00$\times$ & $\infty$ & 1.0000 & 0.0000 & 32.928 & 1.2289 \\
FORA & \textbf{11.621} & \textbf{2.73$\times$} & 12.75 & 0.5093 & 0.5307 & 32.188 & 0.7593 \\
TaylorSeer & 14.651 & 2.17$\times$ & 14.65 & 0.6487 & 0.3973 & 32.505 & 1.1449 \\
HiCache & 16.373 & 1.94$\times$ & 14.93 & 0.6495 & 0.3771 & 32.596 & 1.1638 \\
\textsc{CalibCache} (Ours) & 13.997 & 2.27$\times$ & \textbf{18.18} & \textbf{0.7410} & \textbf{0.2663} & \textbf{32.895} & \textbf{1.1927} \\
\bottomrule
\end{tabular*}
\end{table}
\par\FloatBarrier
\textbf{Timing and implementation details.}
Latency is the mean of 30 CUDA-event measurements after ten benchmark warmup
runs on one NVIDIA H800 per table. It covers the denoising loop and excludes
text encoding, VAE decoding, model loading, calibration, and disk I/O.
For FLUX, the implementation eliminates unconsumed feature-difference work
and uses an output-preserving correction kernel; unused observation work is
also removed from the local Native/FORA measurement wrapper. All 200
CalibCache outputs per grid are pixel-identical to their evaluated outputs.
TaylorSeer, HiCache, and the Qwen methods retain their original implementations.
All timing images match the corresponding evaluated images.
FLUX Native/FORA/CalibCache are measured in interleaved runs within one
resident model; the remaining methods use separate warmed runs on the same
physical GPU within each table.

\textbf{Results across sampling grids.}
Across all six configurations, CalibCache achieves the highest mean
ImageReward, PSNR and SSIM and the lowest mean LPIPS among the evaluated caching
methods. Its denoising speedup exceeds those of TaylorSeer and HiCache in
every configuration, reaching $3.24$--$3.54\times$ on FLUX and
$2.19$--$2.27\times$ on Qwen. These results extend the quality and fidelity
benefits observed at 20 steps to the tested longer schedules.
The reported quality values are 200-prompt means with one fixed noise per
prompt; the comparisons here describe these observed means.

\FloatBarrier